\documentclass[lettersize,journal]{IEEEtran}
\usepackage{amsmath,amsfonts}
\usepackage{array}
\usepackage{textcomp}
\usepackage{stfloats}
\usepackage{url}
\usepackage{verbatim}
\usepackage{graphicx}
\usepackage{cite}
\usepackage[dvipsnames]{xcolor}
\usepackage{paralist}
\usepackage{manyfoot}%
\usepackage{booktabs}%
\usepackage{algpseudocode}%
\usepackage{listings}%
\usepackage{booktabs}
\usepackage{siunitx}
\usepackage{multirow}
\usepackage{diagbox}
\usepackage{xspace}
\usepackage{pifont}
\usepackage{hyperref}

\usepackage[caption=false,font=footnotesize]{subfig}
\makeatletter
\DeclareRobustCommand\onedot{\futurelet\@let@token\@onedot}
\def\@onedot{\ifx\@let@token.\else.\null\fi\xspace}
\newcommand{\revision}[1]{\textcolor{black}{#1}}
\newcommand{\xmark}{\ding{55}}%

\def\ie{\emph{i.e}\onedot}

\makeatother

\begin{document}

\title{Privacy-Preserving Visual Localization with \\Event Cameras}

\author{Junho Kim, Young Min Kim, Ramzi Zahreddine, Weston A. Welge, \\Gurunandan Krishnan, Sizhuo Ma, and Jian Wang
\thanks{The work was done when Junho Kim was an intern at Snap. (\textit{Corresponding authors: Sizhuo Ma and Jian Wang})}
\thanks{Junho Kim and Young Min Kim are with Seoul National University, Seoul, 08826, South Korea (e-mail: 82magnolia@snu.ac.kr; youngmin.kim@snu.ac.kr)}
\thanks{Ramzi Zahreddine, Weston A. Welge, Gurunandan Krishnan, Sizhuo Ma, and Jian Wang are with Snap Inc., Santa Monica CA, 90405, USA (e-mail: rzahreddine@snapchat.com, weston.welge@gmail.com, guru@gurukrishnan.com, sma@snapchat.com, jwang4@snapchat.com)}}

\markboth{IEEE Transactions on Image Processing}%
{Shell \MakeLowercase{\textit{Kim et al.}}: Privacy-Preserving Visual Localization with Event Cameras}

\IEEEpubid{1941-0042~\copyright~2025 IEEE}

\maketitle

\begin{abstract}
We consider the problem of client-server localization, where edge device users communicate visual data with the service provider for locating oneself against a pre-built 3D map.
This localization paradigm is a crucial component for location-based services in AR/VR or mobile applications, as it is not trivial to store large-scale 3D maps and process fast localization on resource-limited edge devices.
Nevertheless, conventional client-server localization systems possess numerous challenges in computational efficiency, robustness, and privacy-preservation during data transmission.
Our work aims to jointly solve these challenges with a localization pipeline based on event cameras.
By using event cameras, our system consumes low energy and maintains small memory bandwidth.
Then during localization, we propose applying event-to-image conversion and leverage mature image-based localization, which achieves robustness even in low-light or fast-moving scenes.
To further enhance privacy protection, we introduce privacy protection techniques at two levels.
Network level protection aims to hide the entire user's view in private scenes using a novel split inference approach, while sensor level protection aims to hide sensitive user details such as faces with light-weight filtering.
Both methods involve small client-side computation and localization performance loss, while significantly mitigating the feeling of insecurity as revealed in our user study.
We thus project our method to serve as a building block for practical location-based services using event cameras.
Project page including the code is available through this link: \url{https://82magnolia.github.io/event\_localization/}.
\end{abstract}


\section{Introduction}
\label{sec:intro}
\begin{figure}[t]
  \centering
  \includegraphics[width=0.9\linewidth]{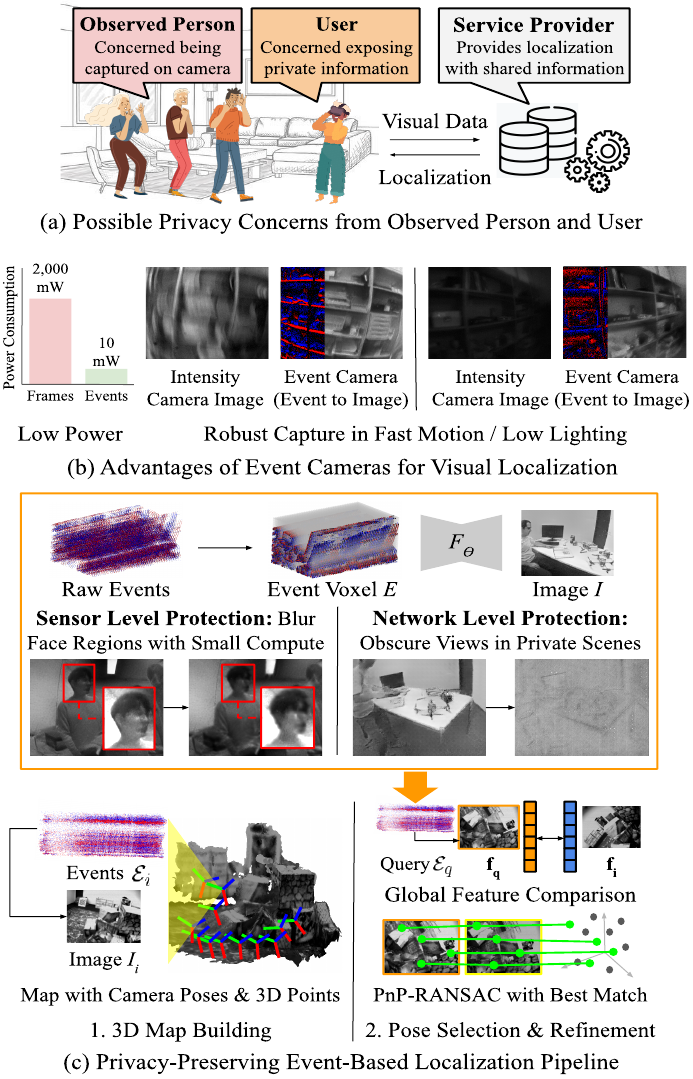}
\caption{
Overview of our approach. \textbf{(a)} Client-server localization introduces privacy concerns. \textbf{(b)} Event cameras have numerous hardware benefits for localization. \textbf{(c)} We achieve privacy-preserving localization by applying protection techniques tailored to events during events pre-processing (sensor level) and event-to-image conversion (network level) (top), where the results are then used for localization (bottom).
}
   \label{fig:teaser}
\end{figure}

\IEEEPARstart{V}{isual} localization is a versatile localization method widely used in AR/VR, which aims to find the camera pose with respect to a pre-built 3D map solely using images.
Due to the limited amount of compute and storage available in AR/VR devices, conventional systems employ \textit{client-server localization} where edge device (\textit{e.g.} smartphones, AR glasses) users transmit visual information to the service provider for localization~\cite{privacy_affine,privacy_sfm,ninjadesc}.
While this enables effective camera pose estimation by reducing the user-side compute burden, privacy concerns arise due to the nature of image capture~\cite{privacy_analysis, privacy_concern}.
As shown in Figure~\ref{fig:teaser}-(a), users of the localization service may be concerned with sharing the current view with the service provider.
Further concerns can be raised by uninformed people who are observed by the user and captured in the localization process.
Along with privacy concerns, localization systems for mobile devices demand for robust performance in a wide range of conditions including fast motion or dark scenes.

\IEEEpubidadjcol
Event cameras are visual sensors that only record brightness \emph{changes}~\cite{survey,orig_event_camera} as a stream of events, which have the potential to provide robust, efficient, and privacy-preserving visual localization.
As shown in Figure~\ref{fig:teaser}-(b), event cameras have a high temporal resolution and dynamic range~\cite{robust_event_tracking}, which are crucial for robust localization in challenging scenarios such as low lighting or fast camera motion.
Further, as the power consumption is far lower than normal cameras~\cite{survey} and the form factor is becoming smaller due to recent advancements in manufacturing~\cite{event_eye,prophesee_eye,prophesee_sensor,sony}, these sensors are highly favorable for machine vision tasks in AR/VR.
From a privacy perspective, the sensors only capture a fraction of visual information as shown in Figure~\ref{fig:teaser}, and thus an average person cannot confidently identify people solely from events.
Nevertheless, this comes at the expense of relatively unstable visual features compared to normal cameras, making direct localization from events difficult.

We propose an event-based visual localization method that can perform robust localization while preserving privacy.
For localization, we employ event-to-image conversion which allows leveraging mature, powerful image-based localization methods~\cite{sarlin2019coarse, robust_retrieval} on captured events.
Our key observation is that despite the information loss during event capture, the converted images contain robust and salient image features sufficient for localization.
The resulting method achieves localization accuracy comparable with an intensity camera in normal scenarios, but significantly better for fast camera motion or low lighting, where localization using normal cameras typically fails.
In addition, by combining event-to-image conversion with image-based localization which effectively reduces the domain gap between events and images, our method can outperform existing event-based localization methods.
To make our solution amenable to mobile devices, the client is only responsible for a light-weight capture and computation process, while the service provider performs the computationally expensive conversion~\cite{event2vid, img_recon_2, img_reconstruction, fast_ev_img, reduce_sim2real, spade_e2vid, hyper_e2vid} and localization steps.


We then design two levels of privacy protection \emph{tailored} for event cameras, as shown in Figure~\ref{fig:teaser}-(c).
On the \textbf{network level}, we observe that naively offloading the computation to the server can lead to privacy breaches. 
To address such concerns, we propose to split neural network inference with a novel re-training procedure for neural networks that prevents the service provider from reconstructing what the users see.
This protection scheme targets users who are willing to use location-based services in private spaces (\textit{e.g.} apartment rooms), where protecting the entire user view should be solicited.
On the \textbf{sensor level}, we propose novel filtering methods based on the spatiotemporal volume of events that preserve important static landmarks for localization, while blurring facial landmarks  without explicit detection.
This process targets wider use cases in both private and public spaces: it aims to reduce concerns about being recorded by wearables or mobile devices.
The filters are designed to be light-weight which makes it possible to implement such protection on the sensor directly.
In practice, since the techniques aim to protect the privacy of different targets (users and observed people), they can be applied jointly.

\begin{figure}[t]
  \centering
  \includegraphics[width=\linewidth]{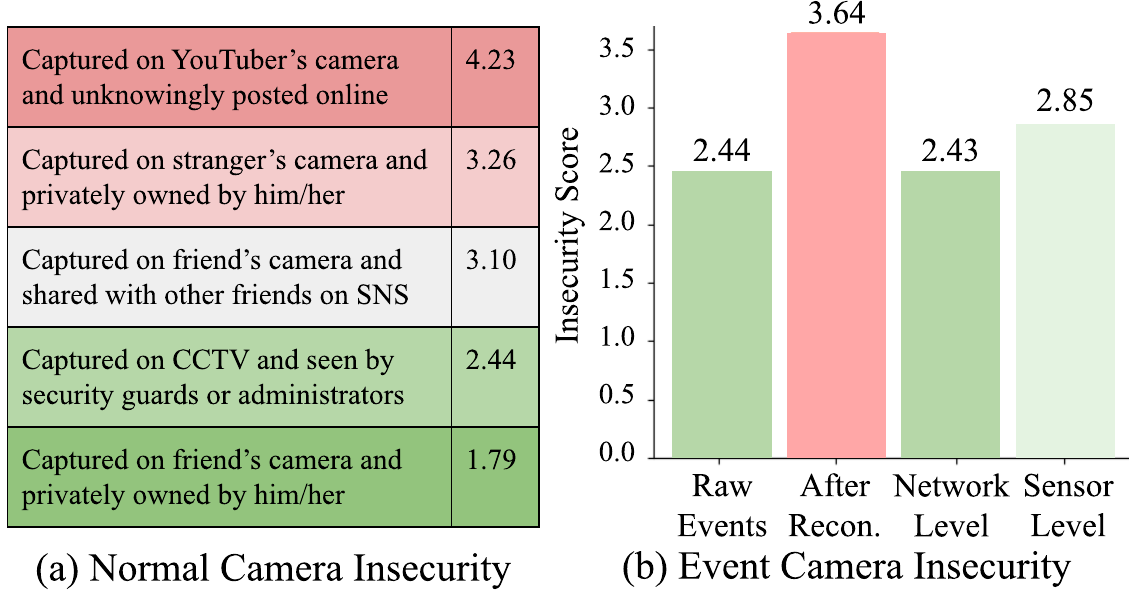}
   \caption{User study results on our privacy protection method. The insecurity scores range between 1 and 5, where higher score indicates higher insecurity. (a) We make an initial measurement on how users feel about being captured using normal cameras in various scenarios. (b) Then, we query on event cameras by sequentially showing raw events, event-to-image reconstructions, and privacy protection results.
   } 

\label{fig:survey}
\end{figure}

We design a rigorous evaluation procedure to assess our method on a wide range of localization scenarios involving moving people, low-lighting, or fast camera motion.
Specifically, we create two new datasets called EvRooms and EvHumans, and introduce protocols for holistically evaluating privacy protection and localization performance.
We also conduct a user study to examine how real users perceive our privacy protection schemes.
Experiments show that our approach effectively balances localization performance with privacy protection.
We first show that, without privacy protection, the proposed event-to-image approach outperforms existing event-based methods that do not rely on image reconstruction~\cite{binary_image_2,lin_evloc,ev_loc_2}.
We then demonstrate that the proposed privacy preservation techniques efficiently protect user privacy without significantly reducing the localization accuracy.
Notably, in the user study, both levels of privacy protection have shown to alleviate privacy concerns of real users, as seen in Figure~\ref{fig:survey}.
We will publicly release the dataset along with the accompanying code for benchmarking, which we expect to foster future research in event-based visual localization.
To summarize, our key contributions are: 
\begin{compactitem}
    \item Proposing the first event-driven effort for robust and private localization targeting mobile environments.
    \item Novel network level privacy protection for mitigating users' concerns.
    \item Novel sensor level privacy protection for relieving observed people's concerns.
\end{compactitem}
By synergistically combining event cameras with robust localization and privacy protection, we expect our method to serve as a practical building block for spatial perception systems in mobile environments.

\section{Related Work}
\label{sec:related}

\subsubsection{Event-Based Visual Localization}

Due to the high dynamic range and small motion blur, event cameras have been widely studied for visual odometry (VO) or SLAM tasks involving \textit{sequential pose estimation} between temporally adjacent events~\cite{ev_slam_hanme_1, ev_slam_hanme_2, ev_slam_representation, ev_sparse_odom, photo_ev_tracking_1, photo_ev_tracking_2,slam_handbook,devo,deep_event_tracking_1,deep_event_tracking_2}.
However, re-localizing an event camera with respect to a pre-built 3D map, namely event-based visual localization, is a fairly understudied problem.
Unlike the VO or SLAM setup, here poses should be estimated amidst large viewpoint differences between the query events and reference images in the 3D map~\cite{sarlin2019coarse,sarlin2020superglue}.
In general, despite recent developments in learning-based feature descriptors for events~\cite{ev_feature_1,ev_feature_2,ev_feature_3,ev_feature_4}, approaches that directly use the raw event measurements usually perform inferior to the counterpart using normal images~\cite{ev_loc_1,slam_handbook,nyu_events}.
This is due to the instability of visual features in events compared to normal images and the lack of large scale training data~\cite{n_imagenet, n_caltech,ev_feat_limit}.
Prior works perform direct camera pose estimation using neural networks~\cite{ev_loc_1, ev_loc_2}, which requires per-scene training and are inferior to structure-based methods in performance~\cite{limit_direct,posenet,lstm_vis_loc}.
Structure-based methods leverage correspondences in 2D and 3D by comparing feature descriptors~\cite{netvlad, robust_retrieval, superpoint, 
sarlin2019coarse, sarlin2020superglue, semantic_local_feats}.
Due to decades of research in image feature descriptors~\cite{sift, superpoint}, structure-based methods are known to perform stable localization.
We leverage event-to-image conversion to combine the mature structure-based paradigm with event cameras, leading to robust localization under challenging conditions.

\subsubsection{Privacy-Preserving Machine Vision}
Machine vision applications in networked environments are subject to privacy breaches as they utilize images captured from the clients~\cite{privacy_analysis, privacy_concern, privacy_sift}.
Recent works target privacy protection in split inference scenarios where the costly neural network computation is shared between the client and server~\cite{privacy_action, privacy_encrypt_face, privacy_freq_domain,split_1,split_2,split_3,split_4,split_5,split_6,split_7}.
Such attempts can be viewed as instances of a broader problem in cryptography called Secure Multiparty Computation (SMC)~\cite{smc}, where the goal is to find methods for parties to compute a function (in this case a neural network) over their inputs while hiding the input values from each other.
Unlike prior works that mainly target classification tasks where a reduced amount of information is sent to the server, we tackle privacy-preservation for reconstruction which requires sending all the information to reconstruct a full image.
This is a more challenging and complex task as it additionally demands hiding the \textit{outputs} of the computation (i.e., reconstructed images) from the server, which differs from the standard SMC setup that mainly focuses on hiding the inputs.

Prior works in privacy-preserving visual localization also focus on this aspect, where existing methods suggest lifting the 2D / 3D keypoints to lines~\cite{privacy_line2d, privacy_line3d, privacy_sfm,coord_perm,obfuscation_3d}, or training a new set of feature descriptors hiding sensitive details~\cite{ninjadesc, privacy_analysis, privacy_affine,gomatch,segloc}.
Another direction of recent works try to encrypt the visual data in the sensor level by incorporating specially designed optics~\cite{privacy_sensing_1, privacy_sensing_2, privacy_sensing_3, privacy_sensing_4}.
For example ~\cite{privacy_sensing_1} designs a phase mask applicable on the camera lens that hides the sensitive scene content and trains a neural network that decodes the phase mask outputs to get depth map predictions.
Inspired by such works, we propose network level privacy protection that prevents possible server-side attacks, and sensor-level protection that hides sensitive visual details in event data.

\section{Event-Based Localization Pipeline}
\label{sec:method}

\label{sec:ev_loc}

Given a short stream of events recorded by an event camera, our method aims to find the 6DoF camera pose within a 3D map as shown in Figure~\ref{fig:teaser}-(c).
Event cameras are visual sensors that track brightness changes as a stream of events, $\mathcal{E}=\{e_i=(x_i, y_i, t_i, p_i)\}$, where $e_i$ indicates the brightness change of polarity $p_i \in \{+1, -1\}$ at pixel location $(x_i, y_i)$ and timestamp $t_i$.
Compared to conventional image sensors, the outputs are sparse and asynchronous, and thus consumes less memory bandwidth and energy which makes the sensors amenable to mobile vision applications.

Our localization method combines event-to-image conversion with image-based localization.
We incorporate such a design choice as our method tackles the previously under-explored problem of finding the event camera pose with respect to a pre-built 3D map, also known as \textit{visual re-localization~\cite{lessmore}}. 
\revision{The problem is challenging due to large viewpoint differences compared to SLAM or odometry scenarios~\cite{devo,deep_event_tracking_1,deep_event_tracking_2}.
As a result, existing methods directly using events~\cite{ev_loc_1, ev_loc_2,ev_feature_1,ev_feature_2,ev_feature_3,ev_feature_4} often exhibit unstable re-localization performance.
Our approach can perform highly accurate localization even in difficult scenarios (fast motion, low lighting) for image-based methods~\cite{superpoint, sarlin2020superglue} by jointly leveraging the sensor-level benefits of events and the maturity of image-based localization.}

Given an input event stream $\mathcal{E}$, let $E$ denote the event voxel grid ~\cite{event2vid,fast_ev_img,unsup_opt_flow} obtained by taking weighted sums of event polarities within spatio-temporal bins.
Event-to-image conversion methods~\cite{event2vid, fast_ev_img,reduce_sim2real,img_recon_1,img_recon_2} take the event voxels as input and produce images using neural networks, namely $F_\Theta(E) = I$ where $\Theta$ denotes the neural network parameters.


\subsection{Structure-Based Localization}
Figure~\ref{fig:teaser}-(c) shows our localization process.
In the mapping stage, a 3D map representaion is built from event streams captured  for a scene: $S_e{=}\{\mathcal{E}_1,\dots,\mathcal{E}_N\}$, with each stream being a spatiotemporal volume of events spanning a short time. In the query stage, a user-captured query stream $\mathcal{E}_q$ is matched against the map to localize the user. To build the map, we first convert the reference events $S_e$ into images $S_i{=}\{I_1,\dots,I_N\}$.
Then we run an off-the-shelf structure-from-motion pipeline COLMAP~\cite{colmap_1} on $S_i$.
The result is a map containing 3D keypoints and 6DoF pose-annotated reference images.

Next, we extract global features vectors using NetVLAD~\cite{netvlad} for each pose-annotated reference image $I_i$ in the 3D map for \emph{candidate pose selection}.
In the query stage, we reconstruct an image $I_q$ from the query event stream $\mathcal{E}_q$ and compute the L2 distances between the query and reference image features. 
We select the top-K nearest poses which serve as candidate poses for further refinement.


Finally, we refine the pose by first performing local feature matching~\cite{sarlin2020superglue,superpoint} between the query and selected reference images.
We count the number of matches found for each query-reference pair and choose the reference view $I_r$ with the largest number of matches.
Then we obtain the refined 6DoF pose by retrieving the 3D points visible from $I_r$ and performing PnP-RANSAC~\cite{pnp_1,ransac} between the 2D points in $I_q$ and retrieved 3D points.

By leveraging event-to-image conversion, we can effectively deploy powerful image-based localization methods on events.
Nevertheless, for high-quality image recovery the conversion solicits repetitive neural network inferences~\cite{event2vid,reduce_sim2real}, which can be costly for edge devices.
This necessitates the transmission of visual information from edge devices to service providers, where we propose various techniques for preserving privacy in Section~\ref{sec:privacy_preserving}.

\section{Privacy-Preserving Localization Overview}
\label{sec:privacy_preserving}
We propose two levels of privacy protection  during information sharing between the user and service provider: network- and sensor-level protection.
\textbf{Network-level privacy protection} targets localization in private scenes ({\it e.g.} apartments, corporate offices), where the user would want to completely hide what they are looking at.
\textbf{Sensor-level privacy protection} targets a broader range of applications and focuses on hiding non-structural details such as facial regions with small additional computation.
\revision{Note that similar to prior works in privacy-preserving visual localization~\cite{ninjadesc,privacy_affine,privacy_analysis}, we assume an \textit{honest-but-curious} service provider~\cite{privacy_threat_model} that faithfully provides the required computation (e.g., global/local feature extraction for localization) but is curious and may attempt to extract the client's visual information from the shared data.
Therefore, the service provider considered in our work cannot secretly gain access to data that the client has not agreed to share (e.g., raw event data), and as a result, our work focuses on secure information sharing during client-server localization.
}

\section{Network-Level Privacy Protection}
\label{sec:network_level}

\begin{figure}[t]
  \centering
  \includegraphics[width=0.85\linewidth]{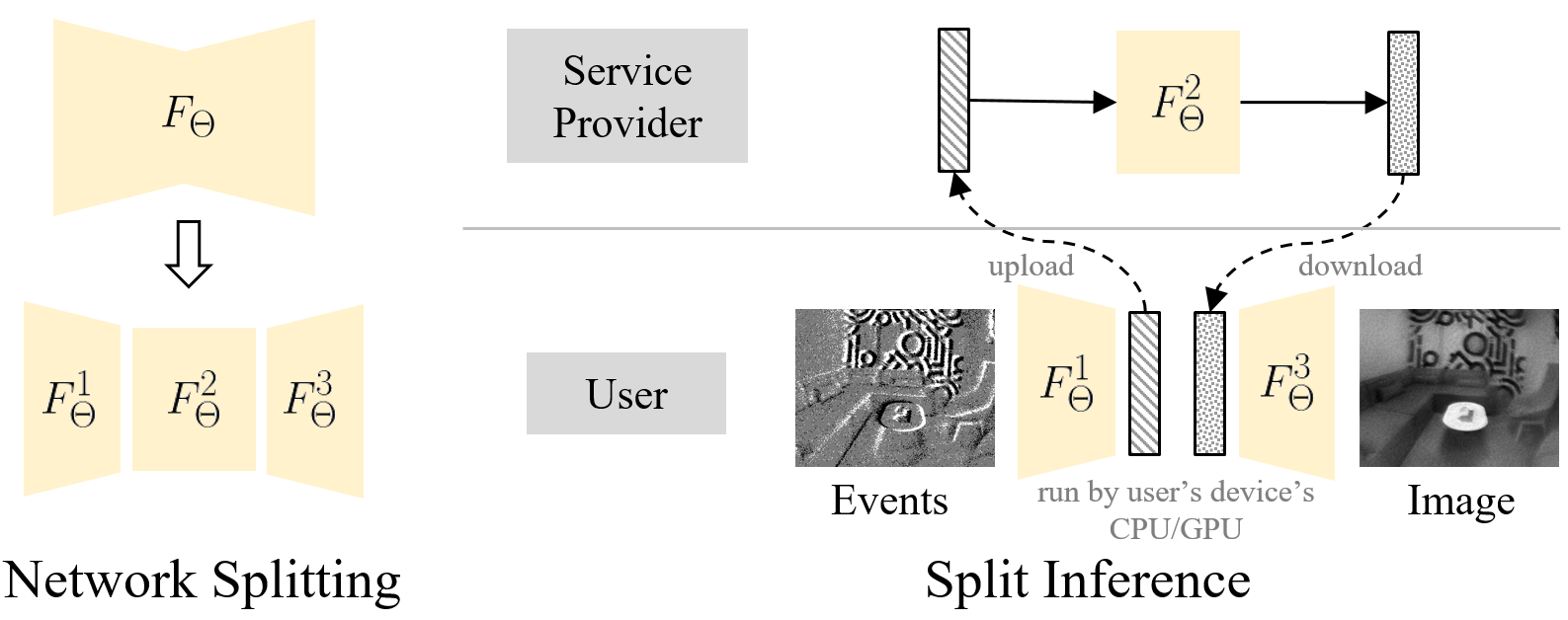}
   \caption{Split inference setup for network level privacy protection. To save compute while hiding sensitive visual information, the original network weights are split (left) and only the costly intermediate part ($F_\Theta^2$) is shared with the service provider. Then, the compute is distributed between the user and service provider: the user performs the light-weight frontal and lateral network inference and the service provider performs the heavy part (right).}
   \label{fig:network_splitted}
\end{figure}

Network level privacy protection hides the user's view from the service provider in private spaces.
As shown in Figure~\ref{fig:network_splitted}, we suggest \textit{splitting} the event-to-image conversion process between the service provider and user, and strategies for preventing possible server-side attacks.
The motivation is twofold: (i) offloading the entire conversion to the server enables revealing the user's view, and (ii) naively splitting inference can still have the server decode the shared intermediate representations.
Below we retain our focus on making the event-to-image conversion process privacy-preserving. Once the images are securely reconstructed, it is also possible to apply existing privacy-preserving visual localization methods~\cite{privacy_affine, privacy_line2d, ninjadesc} for further protection.

\subsection{Split Inference Setup}
\label{sec:splitted}
The split inference process operates in three steps, as summarized in Figure~\ref{fig:network_splitted}.
Prior to inference, the users divide the event-to-image conversion network to disjoint parts $F_{\Theta}^1, F_{\Theta}^2, F_{\Theta}^3$, where $\Theta$ denotes publicly available network weights that are pre-trained on datasets containing general scenes~\cite{coco}. $F_{\Theta}^2$ contains the \textit{majority} of the inference computation and is \textit{the only shared part} with the service provider.
During inference, (i) the user performs inference on $F_{\Theta}^1$, (ii) the result is sent to the service provider to perform $F_{\Theta}^2$, and (iii) the user retrieves the result to finally perform $F_{\Theta}^3$.
Since the frontal and rear inference is done on-device, it may appear at first glance that this conversion is privacy preserving.
However, similar to observations in prior work on split inference~\cite{split_1, split_2}, we identify three possible server-side attacks to decode user information and propose corresponding defense strategies.


\begin{figure}[t]
  \centering
  \includegraphics[width=\linewidth]{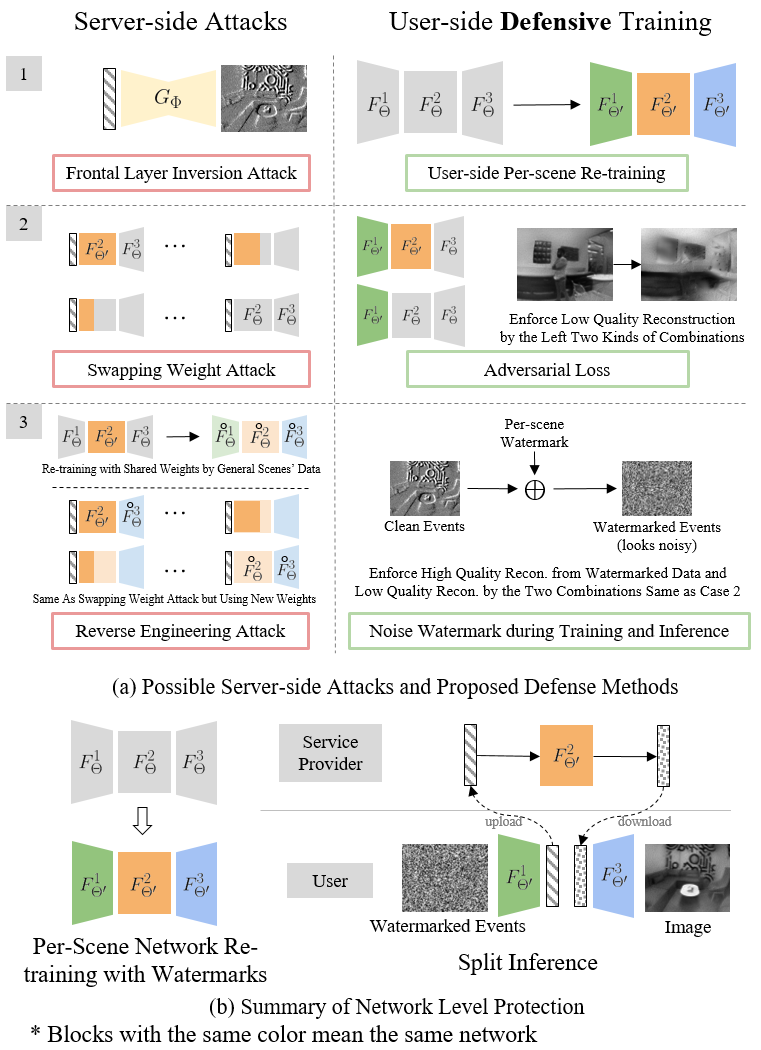}
   \caption{Network level privacy protection targeting users in private scenes. (a) We identify three possible attacks from the service provider. Case 1: Frontal layer inversion attempts to decode the intermediate activations with a learned network $G_\Phi$. Case 2: Swapping weight attack combines the shared network weights with the publicly available weights (gray) to obtain image reconstructions. Case 3: Reverse engineering operates similarly but with network weights trained on the server side aiming to reverse-engineer the unshared network weights. For defense, we propose  per-scene re-training (top) using adversarial losses (middle) and noise-infused event voxels (bottom).
   (b) In the resulting network level protection, users deploy a privately-trained reconstruction network $F_{\Theta^\prime}$ and share the intermediate part with the server during inference.}
   \label{fig:network_level}
\end{figure}

\subsection{Possible Server-Side Attacks and Defenses}
\label{sec:attack_defense}

\subsubsection{Frontal Layer Inversion Attacks}
First, the service provider could train an inversion network $G_\Phi$ that takes the frontal layer activations and regresses the original event voxel.
After that an event-to-image conversion network could be used to obtain an image reconstruction.
If the split inference is performed naively with publicly available pre-trained network weights $\Theta$ \cite{event2vid,spade_e2vid}, then $G_\Phi$ could be easily trained using conventional event camera datasets~\cite{n_caltech, n_imagenet}.
To prevent the attack we propose to have the user to quickly re-train a new event-to-image conversion network $\Theta^\prime$ using events collected in the private scene.
Since the frontal part of the newly trained network $F^1_{\Theta^\prime}$ is unknown to the server, this can prevent training a performant inversion network $G_\Phi$. 
{\bf Note that re-training is a \textit{one-time} operation for each private scene, and can potentially be shared between users residing in the same scene.}
Further, the process does not require large event data as the re-trained network only operates in the private scene.
Typically, re-training can be done using approximately one minute of event camera data, which takes 0.5 hours with a commodity GPU and 3 hours with a CPU.
We provide a detailed analysis on the computational cost of the re-training process in Appendix~\ref{sec:supp_network_eval}.

\subsubsection{Swapping Weight Attacks}
Second, the service provider may use the publicly available network weights $\Theta$ and extract reconstructions from the frontal layer activations.
For example, since the client needs to offload its computation and share the private weights $F^2_{\Theta^\prime}$ to the server, the service provider can combine it with the public weights for the succeeding layers, \ie, run $F_{\Theta}^3 \circ F_{\Theta^\prime}^2$ on top of the shared frontal layer activations $F_{\Theta^\prime}^1(E)$ to reconstruct an image.
Note as in Figure~\ref{fig:network_level}-(a.2), $\Theta^\prime$ and $\Theta$ can be combined at different layers to get an image reconstruction.

To prevent such attacks, we propose to perform re-training with two losses $L{=}L_\text{recon} {+} L_\text{adv}$, where $L_\text{recon}$, $L_\text{adv}$ are the reconstruction and adversarial losses respectively.
Formally, the reconstruction loss is given as follows,
\begin{equation}
    L_\text{recon} = d(F_\Theta(E), F_{\Theta^\prime}({E})),
\label{eq:recon}
\end{equation}
where $d(\cdot, \cdot)$ is the LPIPS distance~\cite{perceptual_loss} and $E$ is the event voxel.
While the reconstruction loss ensures similar image conversion quality to be obtained with the new weights $\Theta^\prime$, the adversarial loss discourages high-quality reconstruction when layers are swapped.
As shown in Figure~\ref{fig:network_level}-(a.2) right half, the loss is defined as the sharpness of the image reconstructions made by swapping parts of the new network layers with the original weights,
\begin{equation}
    L_\text{adv} = s(F_{\Theta}^3 \circ F_{\Theta^\prime}^2 \circ F_{\Theta^\prime}^1(E)) + s(F_{\Theta}^3 \circ F_{\Theta}^2 \circ F_{\Theta^\prime}^1(E)),
\label{eq:adv}
\end{equation}
where $s(\cdot)$ is the gradient magnitude computed by applying Sobel filters~\cite{sobel} on the reconstructions.
Empirically, we found imposing the two losses in Equation~\ref{eq:recon} and ~\ref{eq:adv} to be sufficient for preventing swapping weight attacks.

\subsubsection{Reverse Engineering Attacks}
Finally, the service provider may exploit the shared intermediate weights $F^2_{\Theta^\prime}$ and attempt to \textit{reverse-engineer} the unknown parts of the network.
Specifically, the service provider could perform a re-training on its own using publicly available event data and the loss functions from Equation~\ref{eq:recon} and ~\ref{eq:adv}, but with the intermediate layer weights initialized to $F^2_{\Theta^\prime}$.
After this, the service provider could apply the reverse-engineered weight on the frontal layer activations to obtain an image reconstruction, similar to swapping weight attacks.

For defense we propose to add a noise watermark $E_\text{noise}$ during the client-side re-training process, as shown in Figure~\ref{fig:network_level}-(a.3) right half.
The noise watermark is fixed for each private scene and unknown to the service provider.
The modified training objectives are as follows,
\begin{align}
    \label{eq:recon_noise}
    &L_\text{recon} = d(F_\Theta(E), F_{\Theta^\prime}(\Tilde{E})), \\
    \label{eq:adv_noise}
    &L_\text{adv} = s(F_{\Theta}^3 {\circ} F_{\Theta}^2 {\circ} F_{\Theta^\prime}^1(\Tilde{E})) + s(F_{\Theta}^3 {\circ} F_{\Theta^\prime}^2 {\circ} F_{\Theta^\prime}^1(\Tilde{E})),
\end{align}
where $\Tilde{E}=E + E_\text{noise}$ is the noise-infused event voxel.
We implement the noise watermark $E_\text{noise}$ as a voxel grid with each voxels randomly sampled from the normal distribution $\mathcal{N}(0, 1)$.
During inference, as shown in Figure~\ref{fig:network_level}-(b) the user computes $F_{\Theta^\prime}^1$ from the noise-infused voxel grid $\Tilde{E}$ and sends it to the server, following the steps from Section~\ref{sec:splitted}.
As the noise watermark is unknown to the service provider, it obfuscates the results from user-side re-training and makes reverse engineering attacks to fail.
We validate the effectiveness of network level privacy protection in Section~\ref{sec:network_eval}.

\begin{figure}[t]
  \centering
  \includegraphics[width=\linewidth]{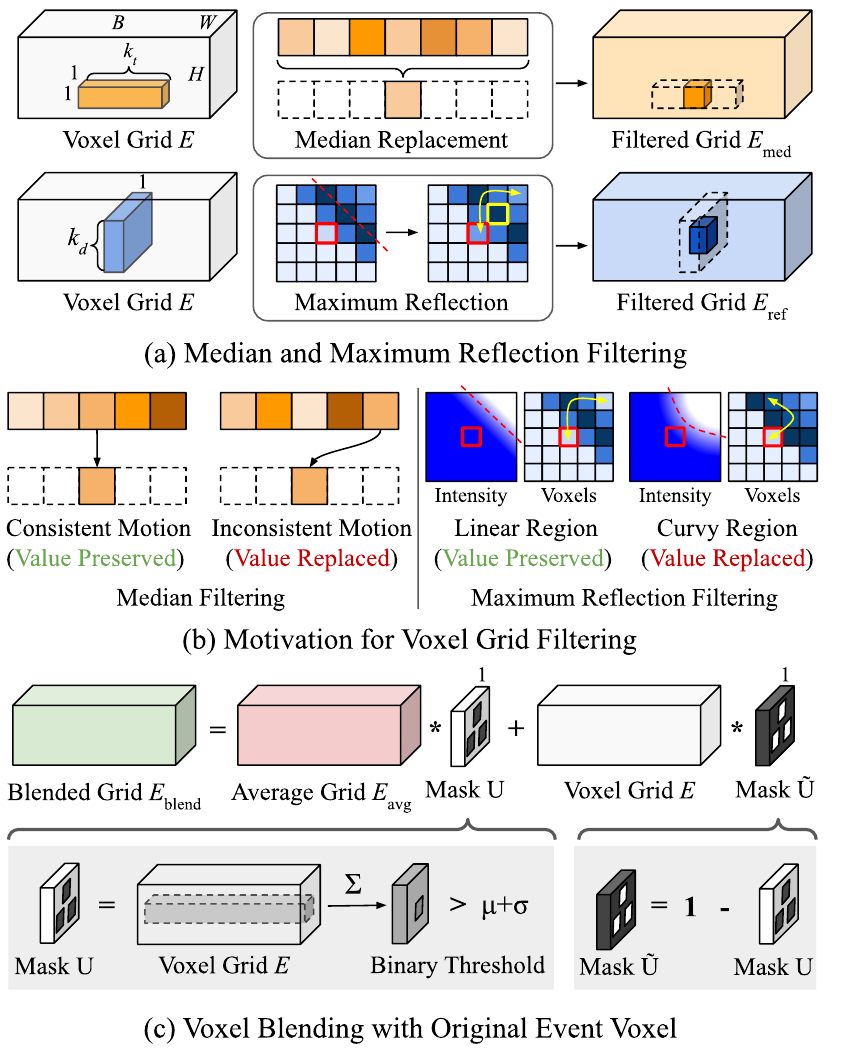}
   \caption{Sensor-level privacy protection. (a) We attenuate temporally inconsistent regions via median filtering and curvy regions via maximum reflection filtering. (b) While the filtering operations can preserve consistent motion or linear regions, the operations will deliberately scramble the values at other regions, leading to blurry reconstructions. 
    (c) To reduce artifacts, the averaged voxels $E_\text{avg}{=}(E_\text{med}{+}E_\text{ref})/2$ are selectively blended with the original voxels.
    Here we use a binary mask $U$ that selects averaged voxels $E_\text{avg}$ only when voxel values are over a threshold.
   }
   \label{fig:sensor_level}
\end{figure}

\section{Sensor-Level Privacy Protection}
\label{sec:sensor_level}
Sensor-level privacy protection aims to hide sensitive details of nearby people, such as faces, observed by the user in both private \textit{and} public scenes.
While network-level protection can also hide scene details from the service provider, it does not address the concerns on data being abused by the user locally. 
Sensor-level privacy protection takes the raw event voxels as input and removes temporally inconsistent or curvy regions that are common on faces, while effectively keeping static, straight structure that is important for localization.
Thus the protection scheme can balance privacy protection with localization performance.
Further, this light-weight processing can potentially be implemented on the sensor directly such that user applications have no access to the raw events.


\subsubsection{Median Filtering}
As shown in Figure~\ref{fig:sensor_level}-(a), median filtering perturbs voxel entries with temporally inconsistent intensity or motion, which are common in events from faces.
Given a voxel grid $E {\in} \mathbb{R}^{B \times H \times W}$ where $B$ denotes the number of temporal bins, we replace each voxel $E(l, m, n)$ with the median value from $E(l {-} k_t{:} l {+} k_t, m, n)$ where $k_t$ is half the temporal window size.
This is based on the intuition that the event accumulation from regions with constant appearance either \emph{monotonically} increase or decrease, where a detailed exposition is given in Appendix~\ref{app:proof}.
On the other hand, dynamic entities including human faces deform over time and the resulting voxel regions show irregularities in the temporal domain, which lead to low quality image reconstructions after filtering.

\subsubsection{Maximum-Reflection Filtering}
We further propose maximum-reflection filtering to attenuate curvy regions that often correspond to facial landmarks.
For each voxel $E(l, m, n)$ we first find the location $(l, m^*, n^*)$ that attains the maximum event count within the spatial neighborhood $|E(l, m {-} k_s{:} m {+} k_s, n {-} k_s{:} n {+} k_s)|$, where $k_s$ is half the spatial window size.
We then replace $E(l, m, n)$ as the voxel value at the reflected location with respect to $(l, m^*, n^*)$, namely $E(l, 2m^* {-} m, 2n^* {-} n)$.
The maximum-reflection filtering preserves event accumulation near lines while replacing other regions with arbitrary values.
The intuition for this operation is that event accumulations near step functions are \emph{symmetrical} with respect to the local maximum, while those on the two sides of a curved line are asymmetrical, as shown in Figure~\ref{fig:sensor_level}b.
Although lines from real-world scenes are not strictly a step function, we find that in practice the maximum-reflection filtering can well-preserve events near lines while attenuating other regions including faces.

Notice that both median filtering and maximum-reflection filtering are unaware of whether there is a face or not (which can be computationally expensive to detect): they blur pixels that \emph{likely} correspond to a face.
While not explicitly detecting faces, such a design choice enables our method to be readily implemented in the sensor level.
Further, despite trading off image reconstruction quality, sensor level protection incurs only a small localization performance drop.
This is further verified in Section~\ref{sec:sensor_eval}. 

\subsubsection{Voxel Blending}
For voxel grid regions with an insufficient amount of accumulations, the filtering process can incur artifacts as the signal-to-noise ratio is low.
Therefore, we blend the filtered voxels with the original event voxel using binary thresholding as depicted in Figure~\ref{fig:sensor_level}-(c).
To elaborate, the binary mask $U \in \mathbb{R}^{B \times H \times W}$ is defined as follows,
\begin{equation}
\label{eq:binary_mask}
    U(l, m, n) = 
\left\{
	\begin{array}{ll}
		1  & \mbox{if \ $E_s(m, n) > \mu + \sigma$} \\
		0 & \mbox{otherwise,}
	\end{array}
\right.
\end{equation}
where $\mu, \sigma$ is the mean and standard deviation of the temporally-summed event accumulations, $E_s(m, n) = \sum_{l}|E(l, m, n)|$.
Then, the blended voxel is given as
\begin{equation}
    E_\text{blend} = U \cdot (\frac{E_\text{med} + E_\text{max}}{2}) + (1 - U) \cdot E,
\end{equation}
where $E_\text{med}, E_\text{max}$ denote the median and maximum-reflection filtered voxels respectively.
The resulting blended voxel $E_\text{blend}$ is then fed to the event-to-image conversion network following the localization pipeline described in Section~\ref{sec:ev_loc}.

\begin{figure}[t]
  \centering
  \includegraphics[width=\linewidth]
  {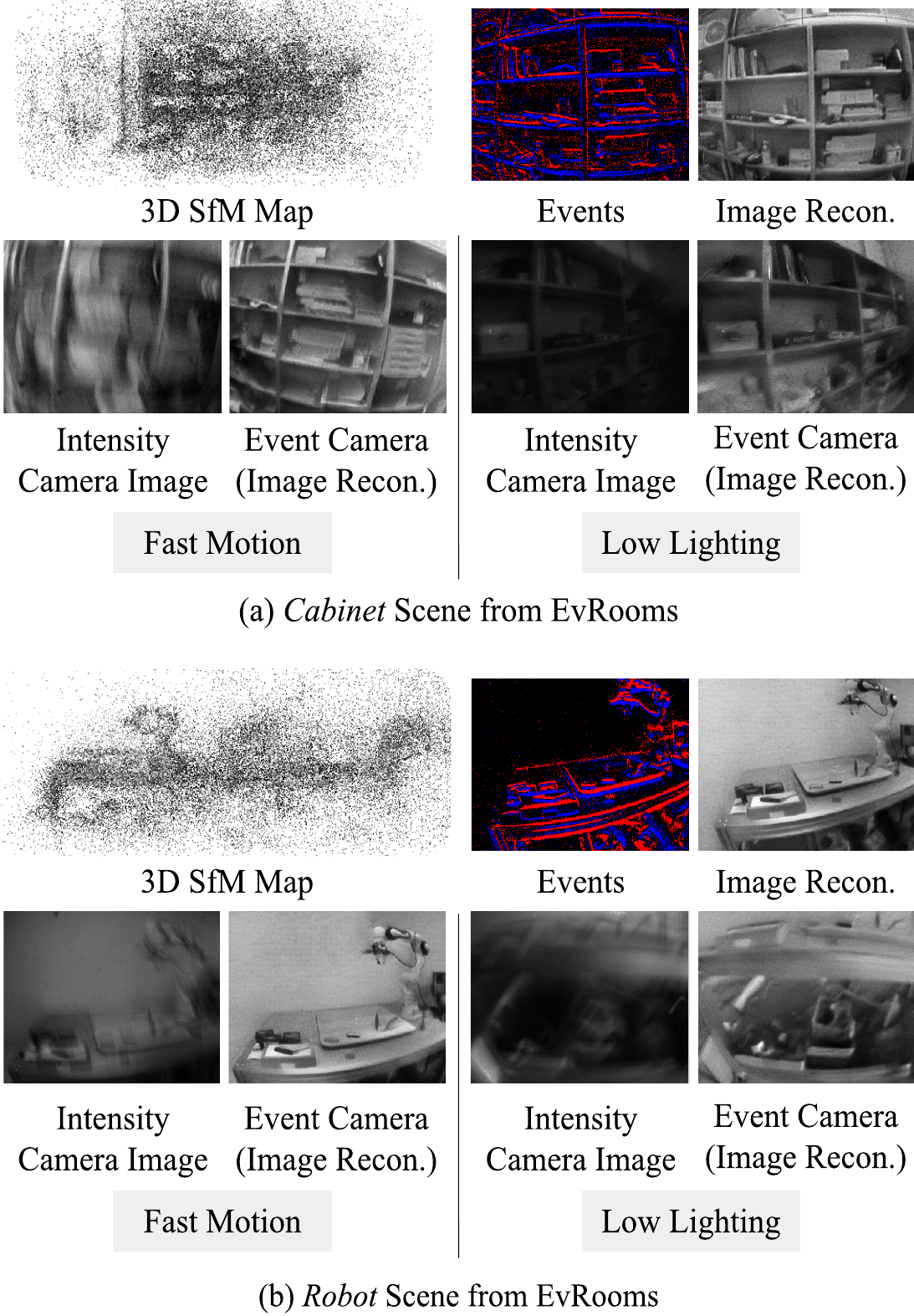}
   \caption{Visualization of 3D maps, events, images, and event-to-image conversions from EvRooms. While the dataset contains challenging scenarios such as fast camera motion or low lighting, event cameras offer stable visual cues for robust localization.}
   \label{fig:dataset}
\end{figure}

\section{Experiments}
We first share user study results on our event-based localization pipeline (Section~\ref{sec:user_study}), then validate the performance of our approach using event-to-image conversion (Section~\ref{sec:loc_eval}), and finally show the quantitative and qualitative analysis of privacy protection (Sections~\ref{sec:network_eval}, ~\ref{sec:sensor_eval}). 
We use three datasets for evaluation: DAVIS240C~\cite{davis_240c}, EvRooms, and EvHumans.
EvRooms is a newly collected dataset to evaluate the robustness of event-based localization algorithms amidst challenging external conditions.
The dataset is captured in 20 scenes and divided into recordings containing fast camera motion (EvRooms\textsuperscript{F}) and low lighting (EvRooms\textsuperscript{L}), where qualitative samples of the 3D structure-from-motion (SfM) maps and images are shown in Figure~\ref{fig:dataset}.
EvHumans is another newly collected dataset for evaluating privacy-preserving localization amidst moving people.
The dataset is captured with 22 volunteers moving in 12 scenes.
All three datasets are captured using the DAVIS346~\cite{davis_346} camera.

Furthermore, we use an RTX 2080 GPU and an Intel Core i7-7500U CPU.
For event-to-image conversion, we adopt E2VID~\cite{event2vid}, which is a conversion method widely used in event-based vision applications~\cite{event_place,event_calib}.
Unless specified otherwise, we use $K{=}3$ candidate poses for refinement in our localization pipeline from Section~\ref{sec:ev_loc}.
For results reporting accuracy, a prediction is considered correct if the translation error is below 0.1 m and the rotation error is below $5.0^\circ$.
All translation and rotation error values are median values, following ~\cite{sarlin2019coarse}.

\subsection{User Study}
\label{sec:user_study}
Before we conduct a detailed empirical analysis on each component of our localization pipeline, we share user study results to illustrate how people would feel about our approach.
We request 39 volunteers to answer a survey that assesses how insecure people feel about various capturing scenarios, where the insecurity is scored from 1 to 5 with larger scores indicating higher insecurity.
As shown in Figure~\ref{fig:survey}, the survey first makes an assessment of being captured with normal cameras for various situations such as tourist spots and CCTV.
Then the participants are asked about their feeling of insecurity with event cameras after seeing (1) raw event measurements, (2) image reconstructions from events, and (3) image reconstructions after network/sensor level protection.
We share the full survey along with the detailed answers of the subjects in the Appendix~\ref{sec:supp_user_study}.

In the initial assessment from Figure~\ref{fig:survey}-(a) people have varying levels of insecurity depending on the capturing scenario.
The averaged insecurity scores provide a rough translation cue for interpreting the scores obtained for event cameras and privacy protections.
In Figure~\ref{fig:survey}-(b), the subjects first give a low insecurity score when they see the raw events but increase their score once they observe that image reconstruction is possible.
The scores drop after people observe the sensor level protection results, to a level roughly equivalent to `being captured on CCTV / friend's camera'.
Furthermore, the scores are even lower for network level protection, as the scene content is completely hidden.
The results show that our method can indeed alleviate the concerns presented by users of AR/VR services and observed people.

\subsection{Localization Performance Analysis}
\label{sec:loc_eval}

\subsubsection{Event-Based Localization Comparison}
We use the DAVIS240C dataset~\cite{davis_240c} for evaluation, and consider seven baselines: direct methods (PoseNet~\cite{posenet}, SP-LSTM~\cite{ev_loc_1}, AECRN~\cite{lin_evloc}), and structure-based methods taking as input various event representations (binary event image~\cite{binary_image_2}, timestamp image~\cite{timestamp_image}) or light-weight image reconstructions from events (complementary reconstruction filter~\cite{scheerlinck_accv}, linear inverse reconstruction~\cite{zhang_lin_inv}).
Note the light-weight reconstruction methods rely on simple optimization techniques for converting events to images, which entail small computation costs but often exhibit inferior reconstruction quality compared to learning based methods as used in our approach~\cite{spade_e2vid, event2vid}.

Table~\ref{table:loc_eval}-(a) shows the localization results of our method and the baselines.
All structure-based methods outperform direct methods, as the pose refinement step using PnP-RANSAC~\cite{epnp, ransac} enables  accurate localization.
Among the structure-based methods, our method outperforms the event representation baselines \cite{binary_image_2,timestamp_image} as the event-to-image conversion mitigates the domain gap and enables leveraging stable image feature descriptors~\cite{netvlad, superpoint}.
A similar trend is present when comparing against light-weight image reconstruction baselines \cite{scheerlinck_accv,zhang_lin_inv}, where the low fidelity of these methods hinders visual feature matching, ultimately deteriorating localization performance.
Therefore, our design choice of incorporating event-to-image conversion is crucial for robust performance to handle a variety of scenarios including fast camera motion and low lighting.

\begin{table}[t]
\caption{Localization evaluation against existing methods.}

\subfloat[Event-Based Localization Methods Comparison]
{
    \centering
    \resizebox{\linewidth}{!}{
    \begin{tabular}{ll|cccc}
        \toprule
        Method & Description & $t$-error (m) & $R$-error ($^\circ$) & Acc.\\
        \midrule
        Direct & PoseNet~\cite{posenet} & 0.15 & 15.94 & 0.05\\
        & SP-LSTM~\cite{ev_loc_1} & 0.19 & 20.30 & 0.03\\
        & AECRN~\cite{lin_evloc} & 0.15 & 15.16 & 0.05\\
        \midrule
        Structure-Based & Binary Event Image~\cite{binary_image_2} & 0.07 & 3.77 & 0.54\\
        &Timestamp Image~\cite{timestamp_image} & 0.06 & 3.18 & 0.58 \\
        & Complementary Recon. Filter~\cite{scheerlinck_accv} & 0.11 & 6.42 & 0.33 \\
        & Linear Inverse Recon.~\cite{zhang_lin_inv} & 0.14 & 10.39 & 0.22 \\
        \midrule
        Ours & Event-to-Image Conversion & \textbf{0.04} & \textbf{2.29} & \textbf{0.69} \\
        \bottomrule
    \end{tabular}
    }
}

\smallskip

\centering
\subfloat[Event Cam. vs. Intensity Cam., Using Image-Based Localization]{
    \resizebox{0.9\linewidth}{!}{
    \begin{tabular}{ll|cccc}
        \toprule
        Dataset Split & Method & $t$-error (m) & $R$-error ($^\circ$) & Acc.\\
        \midrule
        \texttt{Normal} & Intensity Camera & \textbf{0.04} & \textbf{1.77} & 0.72\\
        & Event Camera & 0.05 & 2.00 & \textbf{0.73} \\
        \midrule
        \texttt{Low Lighting} & Intensity Camera & 0.26 & 10.90 & 0.26\\
        & Event Camera & \textbf{0.05} & \textbf{2.53} & \textbf{0.68}\\
        \midrule
        \texttt{Fast Motion} & Intensity Camera & 0.18 & 6.25 & 0.26\\
        & Event Camera & \textbf{0.05} & \textbf{1.82} & \textbf{0.72}\\
        \bottomrule
    \end{tabular}
    }
}

\smallskip

\centering
\subfloat[Localization Evaluation Including Joint Protection]{
    \centering
    \resizebox{0.8\linewidth}{!}{
    \begin{tabular}{l|cccc}
        \toprule
        Method & $t$-error (m) & $R$-error ($^\circ$) & Acc.\\
        \midrule
        No Protection & \textbf{0.04} & \textbf{2.29} & \textbf{0.69} \\
        Network Level Protection & 0.05 & 2.58 & 0.64 \\
        Sensor Level Protection & 0.05 & 2.50 & 0.66 \\
        Joint Protection & 0.06 & 2.88 & 0.62 \\
        \bottomrule
    \end{tabular}
    }

\vspace{-1.5em}    
}
\label{table:loc_eval}
\end{table}

\subsubsection{Image-Based Localization Comparison}
To motivate our focus on event cameras, we implement an exemplary image-based localization method by replacing the input modality in our pipeline from events to intensity images that are captured as a parallel stream by DAVIS cameras.
We create three splits from DAVIS240C and EvRooms: \texttt{\small normal}, \texttt{\small low lighting}, and \texttt{\small fast motion}, where details on data preparation are deferred to Appendix~\ref{sec:supp_dataset}.

Table~\ref{table:loc_eval}-(b) compares localization  results under various settings.
The performance of the two methods are on par in the \texttt{normal} split, as intensity camera-based localization can confidently extract good global/local features in normal conditions.
However, the performance gap largely increases in the \texttt{\small low lighting}, and \texttt{\small fast motion} splits, as the motion blur and low exposure make feature extraction difficult.
Due to the high dynamic range and temporal resolution of event cameras, our method can perform robust localization even in these challenging conditions.

\subsubsection{Localization with Privacy Protection}
We further evaluate the localization performance while using our privacy protection methods in the DAVIS240C dataset~\cite{davis_240c}.
Note as shown in Figure~\ref{fig:network_level}-(b) network-level protection is activated with the re-trained network for image reconstruction in our localization pipeline.
As shown in Table~\ref{table:loc_eval}-(c), the localization accuracy only drops mildly both for sensor and network level, which proves that our protection schemes can effectively balance secure image reconstruction with stable localization.

While network level protection hides the visual content from the service provider, observed people in private spaces may still feel uncomfortable about being captured and stored on mobile devices.
Table~\ref{table:loc_eval}-(c) shows that only a small performance decrease occurs even when both levels of protection are applied.
Thus our method can address privacy concerns from both users and observed people without significantly sacrificing the utility of localization.

\begin{figure*}[t]
  \centering
  \includegraphics[width=0.9\linewidth]{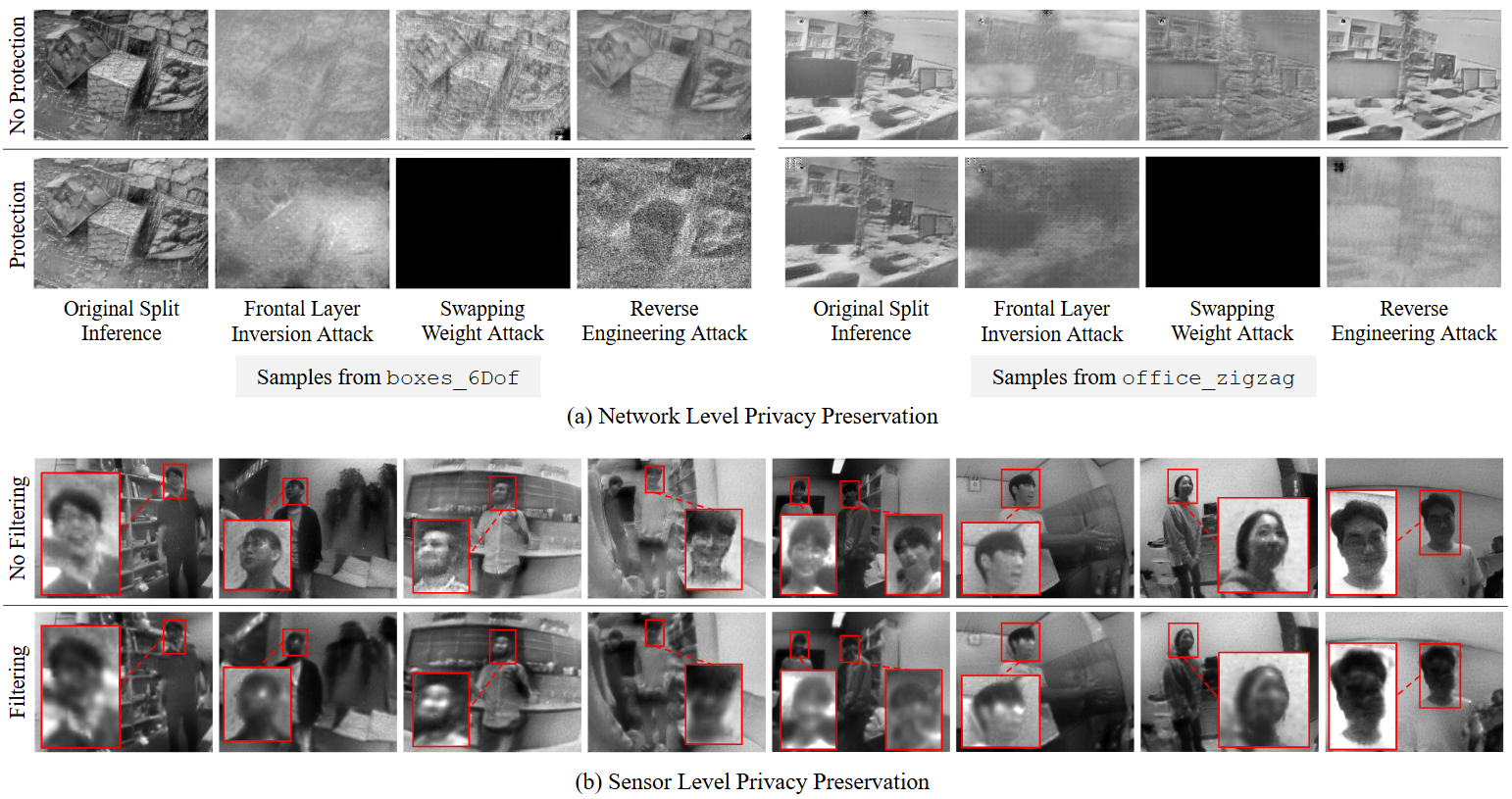}
\caption{Qualitative results of privacy protection. With protection, scene details in (a) and facial information in (b) are significantly removed.}
\label{fig:qualitative}
\vspace{-1.5em}
\end{figure*}

\subsection{Network Level Privacy Preservation Evaluation}
\label{sec:network_eval}
We use six scenes from the DAVIS240C dataset to evaluate network level privacy  protection.
For each scene, we re-train an event-to-image conversion network following Section~\ref{sec:network_level}.
The re-training is quickly done using Adam~\cite{adam} with learning rate $10^{-4}$ and batch size $2$ for 10 epochs.
Then for each trained model, we perform frontal layer inversion and reverse engineering attacks, where the models are trained using events generated from MS-COCO~\cite{coco} following~\cite{event2vid}.
From the E2VID~\cite{event2vid} architecture we use the first two layers as the frontal part ($F_{\Theta^\prime}^1$), the last two layers as the rear part ($F_{\Theta^\prime}^3$), and the rest as the middle part ($F_{\Theta^\prime}^2$).

\subsubsection{Attack Protection Assessment}
We first assess how our method can prevent possible attacks from the service provider.
For each attack, we simulate the procedure from Section~\ref{sec:network_level} by performing image reconstruction where the frontal part of the inference uses the client's network $F_{\Theta^\prime}$ and the latter part uses the service provider's network.
Note that weight swapping is done at various network locations, and we report the averaged image difference between the attacks and the reconstructions from the original network $F_\Theta$.

As shown in Table~\ref{table:network_eval}, the reconstruction quality ({\it i.e.}, deviation from the original network inference) after network level protection is low for all attack scenarios, proving that privacy is protected. 
Simply per-scene re-training with random initialization offers a defense against frontal layer inversion attacks, which can be inferred from large image deviations.
Adversarial loss (Equation~\ref{eq:adv_noise}) plays a key role in blocking swapping weight attacks (`Ours' vs `Ours w/o Adversarial Loss').
For reverse-engineering attacks, a large similarity gap occurs from applying noise watermarking (`Ours' vs `Ours w/o Noise Watermarking') which indicates the crucial role of this procedure for preventing the attack.
We show visualizations of the attacks in Figure~\ref{fig:qualitative}-(a), where all attacks fail after protection.

\subsubsection{Computational Efficiency and Transmission Bandwidth Analysis}
In addition to privacy preservation, network level protection also reduces the computational burden of running the entire event-to-image conversion on-device.
This is similar in spirit to conventional client-server localization systems~\cite{ninjadesc, privacy_affine, privacy_line2d} that perform light-weight compute on the client side and offload majority of the computation to the server.
To assess the computational efficiency, we first compare the event-to-image conversion runtime on CPU, GPU, and our method that performs splitting between the two.
Here the CPU and GPU are used to model the edge device and service provider respectively.
The runtime of our method is 0.08s, which is significantly lower than only using CPU (0.82s), and comparable to the case only using GPU (0.01s).
To further explain such a speedup, we measure the number of floating point operations (FLOPS) for computing each part of the event-to-image conversion network.
Computing intermediate parts of the conversion network ($F_\Theta^2$: 13.82 GFLOPS) is much larger than the frontal and latter parts ($F_\Theta^1$: 0.02 GFLOPS, $F_\Theta^3$: 1.14 GFLOPS), which suggests that computation can be largely reduced by offloading computations of the intermediate network to the server.
While these results may differ from the actual characteristics of edge devices and service providers, our method can efficiently distribute the computation and reduce the burden on the edge device side.

We finally analyze the bandwidth usage of transferring intermediate network outputs.
The average size of the intermediate network outputs to be transferred back and forth between the client and server is 0.75 MB.
This is comparable to the size of the data being transferred in recent split inference works~\cite{split_1} (1.27 MB per network inference), and would consume only a small bandwidth from a commodity 100 MBps WiFi or 20 MBps 4G / 100 MBps 5G cellular network.
Therefore, transferring intermediate network outputs will not be a large bottleneck compared to the actual neural network computation.
Note that in practice, the split network inference only needs to be performed \textit{sparsely} for re-localization and camera tracking after re-localization can be performed using more light-weight event-based methods~\cite{ev_sparse_odom}.

\begin{table}[t]
\caption{Reconstruction Quality (MAE) of Possible Attacks using network level privacy preservation, where per-scene training is performed. 
Note `Vanilla Re-training' refers to using reconstruction loss $L_\text{recon}$ only.}
\centering

    \resizebox{\linewidth}{!}{
    \begin{tabular}{l|cccc}
        \toprule
        \diagbox[width=6cm, height=0.7cm]{Method}{Attack Type} & \begin{tabular}{@{}c@{}}Frontal Layer \\ Inversion\end{tabular} & 
        \begin{tabular}{@{}c@{}}Swapping\\ Weight\end{tabular} &  \begin{tabular}{@{}c@{}}Reverse \\ Engineering\end{tabular}\\
        \midrule
        Vanilla Re-training & 0.3524 & 0.4613  & 0.2165 \\
        Ours w/o Adversarial Loss & 0.4212 & 0.4746 & 0.3517\\
        Ours w/o Noise Watermark & 0.4472 & \textbf{0.9760}  & 0.3237\\
        Ours (Re-training + Adv. Loss + Noise Wtm.) & \textbf{0.4947} & 0.9587  & \textbf{0.4415}\\
        \bottomrule
    \end{tabular}
}

\vspace{-1.5em}
\label{table:network_eval}
\end{table}


\subsection{Sensor Level Privacy Protection Evaluation}
\label{sec:sensor_eval}

We use the EvHumans dataset to assess sensor level protection in face blurring and localization.
In all experiments, we set the half spatial/temporal window size as $k_s{=}23, k_t{=}13$.

\subsubsection{Face Blurring Assessment}
We examine face blurring in terms of low-level image characteristics and high-level semantics.
For evaluation, we generate 9,755 image reconstruction pairs from the event streams with/without sensor level protection.
Also, we use the publicly available FaceNet~\cite{facenet} and DeepFace~\cite{deepface_1} libraries for face detection and description.

Table~\ref{table:sensor_eval} summarizes the evaluation results.
For low-level image analysis, in Table~\ref{table:sensor_eval}-(a) we report the average sharpness of the faces detected from the reconstructed images.
To ensure fair comparison, we first run face detection on the non-filtered image reconstruction and use the detection results to crop both filtered/non-filtered versions. 
The sharpness largely drops after filtering, which indicates that our sensor level protection can effectively blur facial landmarks.
Table~\ref{table:sensor_eval}-(b) further supports this claim, where we measure the image similarity between the two image reconstructions separately for face and background regions. 
The similarity metrics are much higher for background regions, meaning that our method can keep important localization cues ample in the background while blurring out faces.
Some exemplary results are shown in Figure~\ref{fig:qualitative}-(b), where the faces are blurred out from filtering while the background features remain much less intact.

For high-level analysis, Table~\ref{table:sensor_eval}-(a) reports the face detection and grouped face re-identification results.
We apply a face detection algorithm~\cite{mtcnn} on the image reconstructions, where the number of detected faces largely decreases after filtering.
We further analyze how the filtering obfuscates facial features with grouped face re-identification.
In this task, we first divide the faces of volunteers in EvHumans to disjoint groups and apply face re-identification~\cite{arcface} on the detected faces to check whether it belongs to a certain group or not.
Additional details regarding the task are explained in the Appendix~\ref{ref:group_reid_supp}.
Similar to face detection, re-identification accuracy largely drops after filtering, indicating the efficacy of our method to obfuscate facial semantics.

\begin{table}[t]
\caption{Evaluation on sensor level protection.}

\subfloat[Localization and Face Protection Evaluation]{
\centering
\resizebox{0.98\linewidth}{!}{
\begin{tabular}{l|ccc|ccc}
    \toprule
    Method & $t$-error & $R$-error & Acc. & \# of Faces &  Sharpness & Re-ID Acc. \\
    \midrule
    No Protection & \textbf{0.04} & \textbf{0.99} & \textbf{0.84} & 1034 & 0.0956 & 0.9387\\
    w/o Blending & 0.06 & 1.61 & 0.64 & \textbf{106} & \textbf{0.0286} & \textbf{0.4670}  \\
    w/o Max Reflection & \textbf{0.05} & \textbf{1.17} & \textbf{0.77} & 354 & 0.0483 & 0.5708\\
    w/o Median Filtering & \textbf{0.05} & 1.23 & 0.75 & 231 & 0.0461 & 0.5613\\
    \midrule
    Ours & \textbf{0.05} & 1.28 & 0.73 & 192 & 0.0475 & 0.5377\\
    \bottomrule
\end{tabular}
}
}

\centering
\subfloat[Reconstruction Quality of Faces and Background]{
    \centering
    \resizebox{0.7\linewidth}{!}{
    \begin{tabular}{l|cccc}
        \toprule
        Region & PSNR ($\downarrow$) & SSIM ($\downarrow$)& MAE ($\uparrow$) \\
        \midrule
        Face & 18.1620 & 0.3572 & 0.1974 \\
        Background & 21.1243 & 0.6677 & 0.1391 \\
        \bottomrule
    \end{tabular}
    }
}
\vspace{-1.5em}
\label{table:sensor_eval}
\end{table}

\subsubsection{Localization Evaluation and Ablation Study}
We evaluate localization while using sensor level protection, where we pass the filtered voxels to our main localization pipeline.
As shown in Table~\ref{table:sensor_eval}-(a), only a small drop in accuracy occurs.
While attenuating facial features, sensor level protection can preserve important features for localization.

We finally perform an ablation study on the key components of sensor level protection.
As shown in Table~\ref{table:sensor_eval}-(a), using the two filters makes an optimal trade-off between privacy protection and localization performance.
If we ablate the median or max reflection filters, the number of detected faces increases which indicates that the faces are less protected.
However, if we ablate voxel blending, the localization accuracy drastically decreases.
Each component in the sensor level protection is necessary for effective privacy-preserving localization.

\section{Conclusion}
\label{sec:conclusion}
We propose a robust event-based localization algorithm that can simultaneously protect user privacy.
Our method exploits event-to-image conversion to adapt structure-based localization on event cameras for robust localization.
To protect privacy during the conversion, we propose network and sensor level protection.
Network level protection aims to hide the entire view for users in private scenes, whereas sensor level protection targets hiding facial landmarks.
Both levels of protection are light-weight, and our experiments show that the protections incur only small performance drops in localization.
We thus expect our method to be used as a practical pipeline for event-based machine vision systems.

\subsection{Limitations and Future Work}
While we present a first attempt in privacy-preserving event-based localization targeting mobile applications, we acknowledge a number of limitations that solicit future work.
\paragraph{Privacy Protection under a More Capable Adversary}
Our network-level protection assumes an \textit{honest-but-curious} service provider, that only has access to information that the client has agreed to share.
However, our protection scheme may fail for a more capable adversary.
To elaborate, if the service provider gains access to the frontal and lateral neural network weights which are originally kept secretly to the client, then the service provider can run the full inference pipeline to decode the shared neural network activations. 
In addition, if the service provider can access the raw event data captured by the client, it can run event-to-image conversion on the event data to decode the client’s original view.
We expect leveraging recent developments in event data encryption~\cite{event_encrypt_1,event_encrypt_2,event_encrypt_3} to offer protection against such attacks.
\paragraph{Handling Other Biometric Features for Sensor-Level Protection}
\revision{While our sensor-level protection can efficiently blur pixels that likely correspond to a face, we acknowledge that other biometric features such as gait or body shapes may still disclose sensitive information~\cite{ev_gait,gait_recog_1,gait_recog_2,gait_recog_3,body_shape_1}.}
Note this has also been reported by several participants in the user study from Section~\ref{sec:user_study}.
While increasing the level of blurring can hide other biometric features, this will come at the cost of lower localization performance.
Efficient biometric obfuscation methods that preserve the localization performance can be a promising future direction.
\paragraph{Performance Analysis on Real Mobile Hardware}
In the experiments section, we measured various runtime and bandwidth characteristics of our method through software-level simulations.
However, experiments using real mobile hardware will better illustrate the computational requirements of the proposed method.
While such experiments are more difficult to conduct as they require designing and manufacturing real mobile hardware, we acknowledge their importance for accurately measuring real-world feasibility.
\paragraph{Computational Burden for Re-training}
Despite the re-training procedure only requiring small amounts of event camera data, we expect the following directions to further reducing the re-training burden.
First, a single user can retrain the neural network and share it with other users in the private scene.
This way, only one user needs to conduct neural network retraining. Alternatively, multiple users in the private scene may train a single neural network in a distributed manner~\cite{distributed_train_1,distributed_train_2}, which may reduce the training burden for each user.
\paragraph{Hardware Cost of Event Cameras}
\revision{
Despite recent efforts from manufacturers~\cite{prophesee_sensor,sony}, we acknowledge that many event cameras available for purchase today are of relatively high cost compared to conventional RGB cameras.
However, the sensors are increasingly being deployed with a small form factor for mobile setups such as eye tracking in AR/VR devices~\cite{prophesee_eye,event_eye,event_eye_2} or smartphone cameras~\cite{alpsentek}.
Note the pixel size of modern event cameras ($4\sim 5 \mu \text{m}$)~\cite{event_sensor_size} is comparable to that of commodity CMOS image sensors ($1.7 \sim 3.45 \mu \text{m}$)~\cite{cmos_size}.
Finally, we expect that due to economies of scale, larger utilization of these sensors will reduce the overall cost in the future.
}

\section{Acknowledgment}
The authors would like to thank Yicheng Wu for helpful discussions,  volunteers to help construct our EvHumans dataset which is used in our experiment, and volunteers to help our user study on privacy preserving evaluation.


\appendices
\renewcommand\thetable{A.\arabic{table}}    
\setcounter{table}{0}
\renewcommand\thefigure{A.\arabic{figure}}    
\setcounter{figure}{0}
\section{Dataset Details}
\label{sec:supp_dataset}
\subsection{DAVIS240C}
We use six scenes from DAVIS240C~\cite{davis_240c} for localization evaluation in Section~\ref{sec:loc_eval}.
Specifically, we use $\texttt{\small dynamic\_6DoF}$, $\texttt{\small poster\_6DoF}$, $\texttt{\small boxes\_6DoF}$, $\texttt{\small hdr\_boxes}$, $\texttt{\small hdr\_poster}$, and $\texttt{\small office\_zigzag}$ for evaluation.
Among these scenes $\texttt{\small hdr\_boxes}$ and $\texttt{\small hdr\_poster}$ are recorded under low light conditions.

\subsection{EvRooms}
We collect EvRooms using the DAVIS346 camera~\cite{davis_346} from 19 scenes, with 7 scenes additionally recorded under low lighting.
As explained in Section~\ref{sec:ev_loc}, we obtain the 3D maps by first converting short event streams to images~\cite{event2vid} and using an off-the-shelf 3D reconstruction software COLMAP~\cite{colmap_1}.
The dataset contains pose annotations for 18323 images converted from events in fast camera motion (EvRooms\textsuperscript{F}), along with 5022 images for events captured in low lighting (EvRooms\textsuperscript{L}).
For localization evaluation in Section~\ref{sec:loc_eval},  we set the $\texttt{\small Normal}$ split as the four scenes in DAVIS240C ($\texttt{\small dynamic\_6DoF}$, $\texttt{\small poster\_6DoF}$, $\texttt{\small boxes\_6DoF}$, $\texttt{\small office\_zigzag}$).
To evaluate localization in more challenging scenarios, we set the $\texttt{\small Low Lighting}$ split as the two scenes in DAVIS240C ($\texttt{\small hdr\_boxes}$, $\texttt{\small hdr\_poster}$) along with EvRooms\textsuperscript{L}, and the $\texttt{\small Fast Motion}$ split as the scenes in EvRooms\textsuperscript{F}.
We are planning to release the EvRooms dataset in public.

\subsection{EvHumans}
For evaluating localization amidst moving people, we use the newly captured dataset called EvHumans.
Similar to EvRooms, this dataset was captured using the DAVIS346 camera~\cite{davis_346} and we obtain the 3D maps using the event-to-image conversions with COLMAP~\cite{colmap_1}.
The dataset consists of 20 scenes with 16 volunteers, where each scene on average contains $2{\sim}3$ moving people.
Prior to dataset capture, we obtained the consent from all 16 volunteers under the conditions that the dataset is not made public.
We additionally obtained approval from the volunteers whose faces appear in the paper figures.

\section{Sensor Level Protection Details}
\label{sec:supp_sensor_level}
\subsection{Accelerating Filtering}
We further reduce the runtime of sensor level protection by exploiting the fact that the blending process only keeps the filtered values for pixels with sufficient number of event accumulations.
Namely, we sparsely apply the filtering operation only on the pixel regions where the binary mask in Equation~\ref{eq:binary_mask} is non-zero.
This simple optimization reduces the runtime from 4.32 s to 0.15 s for processing $3\times 10^5$ events spanning approximately 0.45 s.
Note that the runtime is measured from filters implemented using pure Python code, and the process could be further accelerated by re-implementing the filters with faster languages such as C++.

\subsection{Justification of Median Filtering}
\label{app:proof}
Among the two voxel filtering steps, median filtering attenuates temporally inconsistent voxel regions.
Here we give a mathematical justification of median filtering based on the event generation equation~\cite{survey}.
First, let $L(\mathbf{u}, t)$ denote the log intensity at pixel $\mathbf{u}$ in time $t$.
Then, assuming constant illumination and motion, we can write the event accumulation from a short time interval $[t - \Delta t, t]$ as follows~\cite{survey},
\begin{equation}
    \Delta L(\mathbf{u}, t) = \nabla L(\mathbf{u}, t) \cdot \mathbf{v} \Delta t,
    \label{eq:event}
\end{equation}
where $\mathbf{v}$ is the apparent velocity at each pixel (optical flow).
Similarly, the event accumulations over a short time $\Delta t$ for the neighboring timestamps $t \pm \Delta t$ could be expressed as $\Delta L(\mathbf{u}, t \pm \Delta t) = \nabla L(\mathbf{u}, t \pm \Delta t) \cdot \mathbf{v} \Delta t$.
Using Taylor expansion and the constant velocity/illumination assumption, we have
\begin{align}
    &L(\mathbf{u}, t \pm \Delta t) = L(\mathbf{u} \mp \mathbf{v}\Delta t, t), \\
    & L(\mathbf{u} \mp \mathbf{v}\Delta t, t) = L(\mathbf{u}, t) \mp \nabla L(\mathbf{u}, t) \cdot \mathbf{v}\Delta t.
    \label{eq:taylor}
\end{align}
By applying Equation~\ref{eq:taylor} on Equation~\ref{eq:event}, we have
\begin{align}
    & \nabla L(\mathbf{u}, t \pm \Delta t) = \nabla L(\mathbf{u}, t) \mp \nabla^2 L(\mathbf{u}, t) \cdot \mathbf{v}\Delta t, \\
    & \Delta L(\mathbf{u}, t \pm \Delta t) = \Delta L(\mathbf{u}, t) \mp \mathbf{v} \cdot \nabla^2 L(\mathbf{u}, t) \cdot \mathbf{v}(\Delta t)^2,
\end{align}
which indicates that in regions with constant velocity and illumination, the event accumulations are either monotone increasing or decreasing for a short period of time.
Therefore, applying median filtering on voxels can preserve the accumulation values for temporally consistent regions while perturbing the values for other regions.

\subsection{Justification of Maximum-Reflection Filtering}
In maximum-reflection filtering we use the fact that pixels near straight edges have image gradient magnitudes $|\nabla L(\mathbf{u}, t)|$ that are line symmetric.
Since the pixel velocity $\textbf{v}$ for static objects are near constant under a moving camera, the event equation in Equation~\ref{eq:event} implies that the event accumulations are nearly symmetrical (Figure~\ref{fig:sensor_level}-(b)).
Thus, events near straight edges are preserved while those at curvy regions are obfuscated.
Although curvy regions do not necessarily correspond to faces, faces are mostly composed of curvy regions. 
Therefore this filtering provides a conservative way to obfuscate faces, at the cost of slight decrease (1 cm, $0.3^\circ$ from Table~\ref{table:sensor_eval}) in localization performance.

\section{Experiment Details}
\subsection{Event Voxels for Image Conversion}
In all our experiments we first package the input events to event voxel grids and apply event-to-image conversion methods~\cite{event2vid,fast_ev_img}.
Given an event stream $\mathcal{E}$ spanning $\Delta T$ seconds, the event voxel is defined by taking weighted sums of event polarities within spatio-temporal bins.
Formally, each entry of the event voxel $E \in \mathbb{R}^{B \times H \times W}$ is given as follows,
\begin{equation}
\label{eq:voxel}
    E(l, m, n) = \sum_{\substack{x_i=l\\ y_i=m}} p_i \max(0, 1 - |n - t^*_i|),
\end{equation}
where $t_i^* = \frac{B-1}{\Delta T} (t_i - t_0)$ is the normalized event timestamp.
For all our experiments, we set the number of temporal bins for the event voxel as $B{=}50$.

\subsection{Localization Evaluation}
\label{sec:supp_loc_eval}

\subsubsection{Query/Reference Split}
We explain the query and reference sets for evaluating localization.
Recall from Section~\ref{sec:ev_loc} that we build the 3D map from events-to-image conversions in the reference set and measure localization accuracy using the query set.
For evaluation in EvRooms and DAVIS240C, we use the first $70\%$ of the event streams for reference and the rest for querying.
On the other hand for EvHumans, we randomly slice $70\%$ of the event streams for reference and the rest for querying.
The distinction is made for EvHumans because the dataset does not contain significant visual overlaps between the frontal and latter events as the capturing was made while constantly tracking the humans close by.

\subsubsection{Baselines}
In Section~\ref{sec:loc_eval} we consider nine baselines for event-based visual localization: direct methods (PoseNet~\cite{posenet}, SP-LSTM~\cite{ev_loc_1}, AECRN~\cite{lin_evloc}), and structure-based methods taking as input various event representations (binary event image~\cite{binary_image_2}, timestamp image~\cite{timestamp_image}) or light-weight image reconstructions from events (Scheerlink et al.~\cite{scheerlinck_accv}, Zhang et al.~\cite{zhang_lin_inv}).
All the direct methods are trained to directly regress the 6DoF pose within a scene, and needs to be re-trained for each scene.
We train the networks separately for each scene in DAVIS240C~\cite{davis_240c}, using the events in the reference split.
Here the pose annotations from the 3D map are used to ensure that the network pose predictions are consistent with our method's prediction.
The networks are all trained for $1400$ epochs using a batch size of 20 and a learning rate of $1e{-}4$ with Adam~\cite{adam}.

The structure-based methods are contrived baselines that either directly package the input events for localization or leverage simple optimization techniques for converting events to images.
Two conventional packaging methods are tested, namely the binary event image caching the pixel-wise event occurrences and timestamp image~\cite{timestamp_image} caching the pixel-wise most recent event timestamps.
In addition, we consider two optimization-based image conversion methods, where Scheerlinck et al.~\cite{scheerlinck_accv} performs numerical integration on event polarities and Zhang et al.~\cite{zhang_lin_inv} performs linear programming for image reconstruction.  
We use the poses from the 3D maps built from event-to-image conversions, but replace the images to corresponding input representations during localization.

\subsection{Network Level Privacy Protection Evaluation}
\label{sec:supp_network_eval}

\subsubsection{Evaluation Protocol}
\begin{figure}[t]
  \centering
  \includegraphics[width=\linewidth]
  {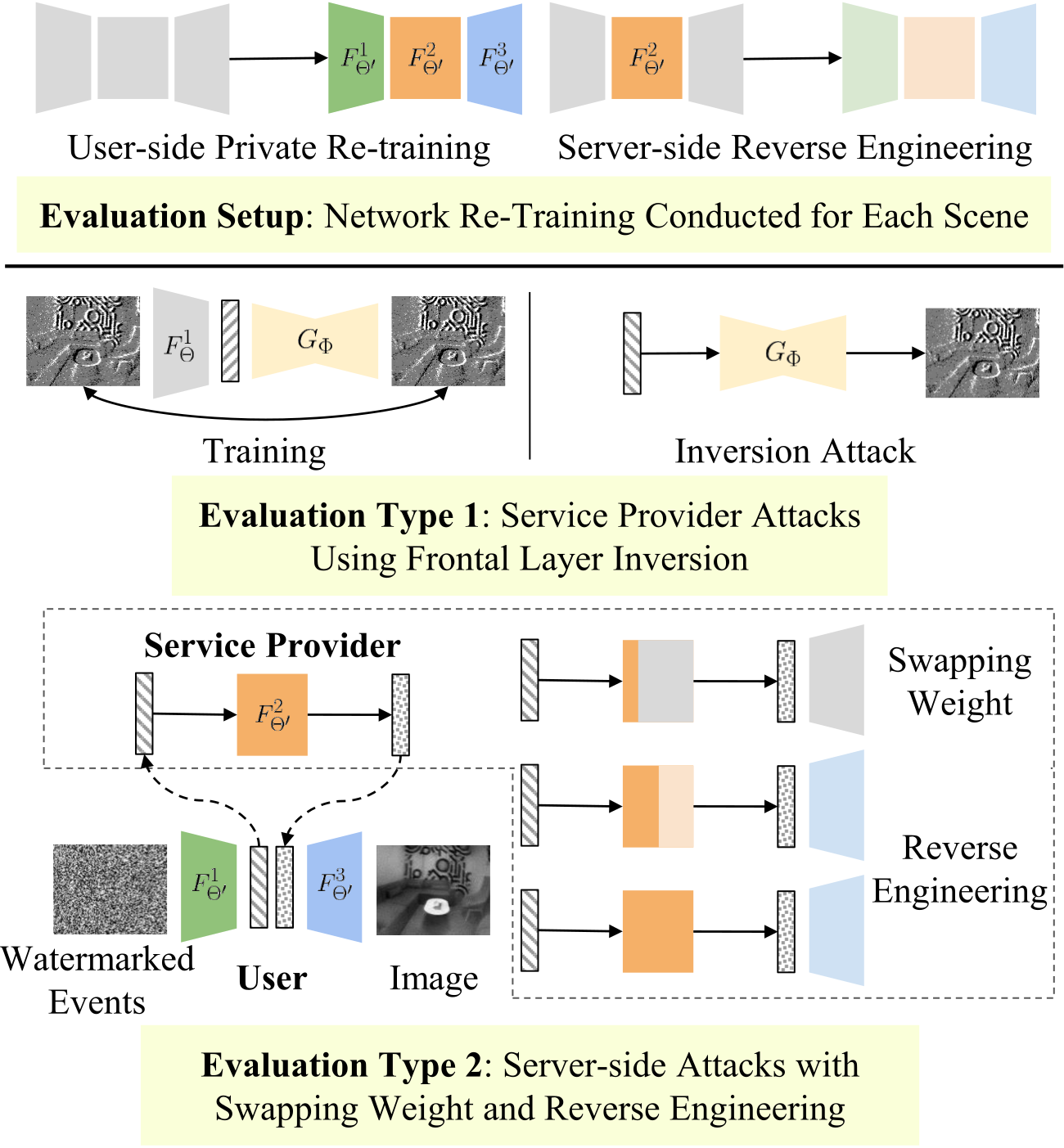}
   \caption{Evaluation procedure of network level privacy protection. \textbf{Evaluation Setup}: For each tested scene, we first train two sets of neural networks: private re-training in the user side (left) and reverse engineering in the service provider side (right). \textbf{Evaluation Type 1}: For frontal layer inversion attack, the service provider trains a network that learns to recover the original event voxel from intermediate layer activations. The recovered voxels are then fed to an event-to-image conversion network for evaluation. \textbf{Evaluation Type 2}: Then, to evaluate swapping weight and reverse engineering attacks we perform inferences using network weights available to the service provider. Specifically, the frontal layer activations (dashed line box) are decoded using the two types of weights available to the service provider at various swapping locations: publicly available weights (gray) for swapping weight attacks and reverse engineered weights for reverse engineering attacks (light orange).}
   \label{fig:network_eval_protocol}
\vspace{-1em}
\end{figure}

We elaborate on evaluating network level privacy protection, where the evaluation process is summarized in Figure~\ref{fig:network_eval_protocol}.
We use the 6 scenes from DAVIS240C~\cite{davis_240c} as in Section~\ref{sec:loc_eval} and first conduct user-side private re-training for each scene (Figure~\ref{fig:network_eval_protocol}: \textbf{Evaluation Setup}).
Here we train each conversion network using the events from the reference set described in Appendix~\ref{sec:supp_loc_eval}.
After this, using the shared weights $F^2_{\Theta^\prime}$, we conduct server-side reverse engineering with general scenes, namely 3,000 event sequences each spanning 2 seconds  generated from the MS-COCO dataset~\cite{coco} following ~\cite{event2vid, fast_ev_img}.
We find that fine-tuning the shared weights improved attack performance compared to freezing them, so we report the results from fine-tuning.
Then, we train the frontal layer inversion network $G_\Phi$ that takes the frontal activations of the publicly available network as input and learns to regress the original event voxel (Figure~\ref{fig:network_eval_protocol}: \textbf{Evaluation Type 1}).
We use the UNet architecture~\cite{unet} to implement $G_\Phi$ and use events from MS-COCO dataset~\cite{coco}.
Finally, to make attacks for each scene we swap the weights during split inference using the original network weights $\Theta$ or the reverse engineering weights $\Theta^\prime$ (Figure~\ref{fig:network_eval_protocol}: \textbf{Evaluation Type 2}).
Note for the frontal layer inversion attack we take the shared intermediate activation and apply the inversion network $G_\Phi$ to recover the original event voxel.
The result is then fed to the reconstruction network $F_\Theta$ to obtain the recovered image.

\subsubsection{Network Re-training Procedure}
We describe the detailed steps on re-training neural networks in the user side for network-level protection.
First, the user residing in a private scene collects a small set of event camera data (approximately spanning 1 minute) and converts the raw events to voxels for re-training following Equation~\ref{eq:voxel}.
Then, the user trains the randomly initialized reconstruction network using the training objectives from Equation~\ref{eq:recon_noise} and \ref{eq:adv_noise} which includes the noise watermark added to each training sample.
Finally, the user shares the intermediate part of the trained network with the service provider.
During inference, given a stream of events, the user first performs inference with the frontal layer and sends the results to the service provider (Figure~\ref{fig:network_eval_protocol}: \textbf{Evaluation Type 2}).
The service provider then performs the intermediate layer inference and returns the result to the user.
The user finally performs the lateral layer inference to obtain the image reconstruction.

\begin{table}[t]
\caption{
Computation cost comparison of re-training E2VID~\cite{event2vid} for privacy protection against the original setup and PoseNet~\cite{posenet}, a learning-based event localization method. The training time for E2VID (Original) is omitted as the training code is not publicly available.
}
\centering
\resizebox{0.8\linewidth}{!}{
\begin{tabular}{l|ccc}
    \toprule
    Method & \begin{tabular}{@{}c@{}} E2VID~\cite{event2vid} \\ (Original) \end{tabular} & \begin{tabular}{@{}c@{}} E2VID~\cite{event2vid} \\ (Private) \end{tabular} & PoseNet~\cite{posenet} \\
    \midrule
    Event Duration (s) & 2000.00 & 70.70 & 70.70 \\
    Number of Epochs & 160 & 10 & 1400 \\
    Training Time (hr) & N/A & 0.5 & 48 \\
    \bottomrule
\end{tabular}
}
\label{table:retrain_cost}
\end{table}
\subsubsection{Re-training Overhead Quantification}
We provide a quantification of the re-training process for network-level protection from Section~\ref{sec:network_level}.
Table~\ref{table:retrain_cost} shows the training data characteristics (average duration of event data, number of epochs, and training time) of our retraining scheme compared against the original training setup of E2VID~\cite{event2vid} and PoseNet~\cite{posenet}, a learning-based localization algorithm.
Our retraining is done for 10 epochs in all experiments, which takes on average 0.5 hours using an RTX 2080 GPU and 3 hours using an Intel Core i7-7500U CPU.
\revision{Further, the training process consumes on average 178.63 W on the GPU and 52.75 W on the CPU.}
In addition, the amount of event data used for training our event-to-image conversion network is much smaller than the original setup for training E2VID: approximately 1 minute of event camera data is required.
Therefore, our retraining scheme can operate quickly compared to other methods.
Further, note that the retraining process is a one-time operation for each private scene: once the network is retrained, it can be continuously deployed within the same scene for visual localization.

\subsection{Sensor Level Privacy Protection Evaluation}
\label{sec:supp_sensor_eval}

\subsubsection{Group Re-Identification Evaluation Protocol}
\label{ref:group_reid_supp}
We elaborate on the details of the grouped face re-identification task from Table~\ref{table:sensor_eval} and Section~\ref{sec:sensor_eval}.
We first divide the faces of volunteers in EvHumans to disjoint groups and apply face re-identification~\cite{arcface} on the detected faces to check whether it belongs to a certain group or not.
To elaborate, we first run face detection~\cite{facenet} on the non-filtered reconstructed images  and obtain the bounding boxes for the detected faces.
We then use the bounding boxes to crop facial regions in the filtered image reconstructions similar to the image sharpness evaluation.

For each scene, we make two groups $\mathcal{G}_\text{in}$ containing the faces of all the people appearing in the scene and $\mathcal{G}_\text{out}$ containing the faces of the remaining people.
Then we further split $\mathcal{G}_\text{in}$ into two groups $\mathcal{G}_\text{in}^\text{ref}, \mathcal{G}_\text{in}^\text{test}$ with $\mathcal{G}_\text{in}^\text{ref}$ containing the first $80\%$ of the faces and $\mathcal{G}_\text{in}^\text{test}$ containing the rest.
Finally, for each face in $\mathcal{G}_\text{in}^\text{test}$ we (i) extract the ArcFace~\cite{arcface} descriptor and compare against the descriptors extracted for faces in $\mathcal{G}_\text{in}^\text{ref}$ and $\mathcal{G}_\text{out}$, and (ii) choose the group with the smallest L2 descriptor distance.
We finally measure the ratio of predictions with the correct group re-identifications.

\begin{table}[t]
\caption{Privacy protection comparison against simple filtering methods, namely Gaussian and mean filtering. 
}
\centering

\resizebox{\linewidth}{!}{
\begin{tabular}{l|ccc|ccc}
    \toprule
    Method & $t$-error & $R$-error & Acc. & \# of Faces &  Sharpness & Re-ID Acc. \\
    \midrule
    No Filtering & \textbf{0.04} & \textbf{0.99} & \textbf{0.84} & 1034 & 0.0956 & 0.9387\\
    Gaussian Filtering & 0.11 & 2.85 & 0.42 & \textbf{55} & \textbf{0.0341} & \textbf{0.4151}\\
    Mean Filtering & \textbf{0.04} & 1.12 & 0.77 & 372 & 0.0449 & 0.5755\\
    \midrule
    Ours & 0.05 & 1.28 & 0.73 & 192 & 0.0475 & 0.5377\\
    \bottomrule
\end{tabular}
}
\vspace{-1em}
\label{table:supp_filter}
\end{table}

\subsubsection{Comparison against Gaussian and Mean Filtering}
We make comparisons against more simpler design choices for voxel filtering, namely Gaussian and mean filtering.
The filtering operations are performed with a spatio-temporal kernel size of $5 {\times} 5 {\times} 5$, and the blending operations is kept identically as our sensor level protection method.
Note the hyperparameters for filtering were chosen to attain the optimal balance between privacy and localization performance.
The results are summarized in Table~\ref{table:supp_filter}.
Gaussian filtering results in harsher removal of feature points and incurs large drop in localization accuracy.
Mean filtering on the other hand shows smaller localization performance drop but fails in protecting facial information which could be indicated from the large number of face detections.
Our method better balances between privacy protection and localization performance preservation.

\subsubsection{Additional Ablation Study}
We conduct an additional ablation study on the effect of various hyperparameters on privacy protection and localization performance of sensor-level protection.
Table A.3 summarizes the number of faces detected, which measures privacy protection, and localization accuracy under variations in filter size and voxel blending. 
We follow the experimental setup from
Section~\ref{sec:sensor_eval} for evaluation.
While the performance of our method does not largely fluctuate by the choice of hyperparameters, our design choice of setting the median filter size $k_t = 13$, maximum-reflection filter size $k_s = 23$, and using voxel blending shows the optimal balance between privacy protection and localization accuracy.
\begin{table}[t]
\label{table:supp_filter_2}
\caption{Evaluation on the effect of various hyperparameters on median (med.) and maximum-reflection (max-ref.) filters for sensor-level protection performance.}
\centering
\resizebox{0.98\linewidth}{!}{
\begin{tabular}{l|ccc|cc}
    \toprule
    Method & \begin{tabular}{@{}c@{}} Med. \\ Filter Size \end{tabular} & \begin{tabular}{@{}c@{}} Max-Ref. \\ Filter Size \end{tabular} & \begin{tabular}{@{}c@{}} Voxel \\ Blending \end{tabular} & \begin{tabular}{@{}c@{}} Loc. \\ Acc. \end{tabular} & \begin{tabular}{@{}c@{}} \# of \\ Faces \end{tabular} \\
    \midrule
    w/o Voxel Blending & 13 & 23 & \xmark & 0.64 & \textbf{106} \\
    w/ Larger Med. Filter & 21 & 23 & $\bigcirc$ & 0.71 & 185 \\
    w/ Smaller Med. Filter & 7 & 23 & $\bigcirc$ & 0.72 & 210 \\
    w/ Larger Max-Ref. Filter & 13 & 31 & $\bigcirc$ & 0.72 & 184 \\
    w/ Smaller Max-Ref. Filter & 13 & 11 & $\bigcirc$ & \textbf{0.75} & 278 \\
    \midrule
    Ours & 13 & 23 & $\bigcirc$ & 0.73 & 192 \\
    \bottomrule
\end{tabular}
}
\end{table}

\begin{figure}[t]
  \centering
  \includegraphics[width=\linewidth]
  {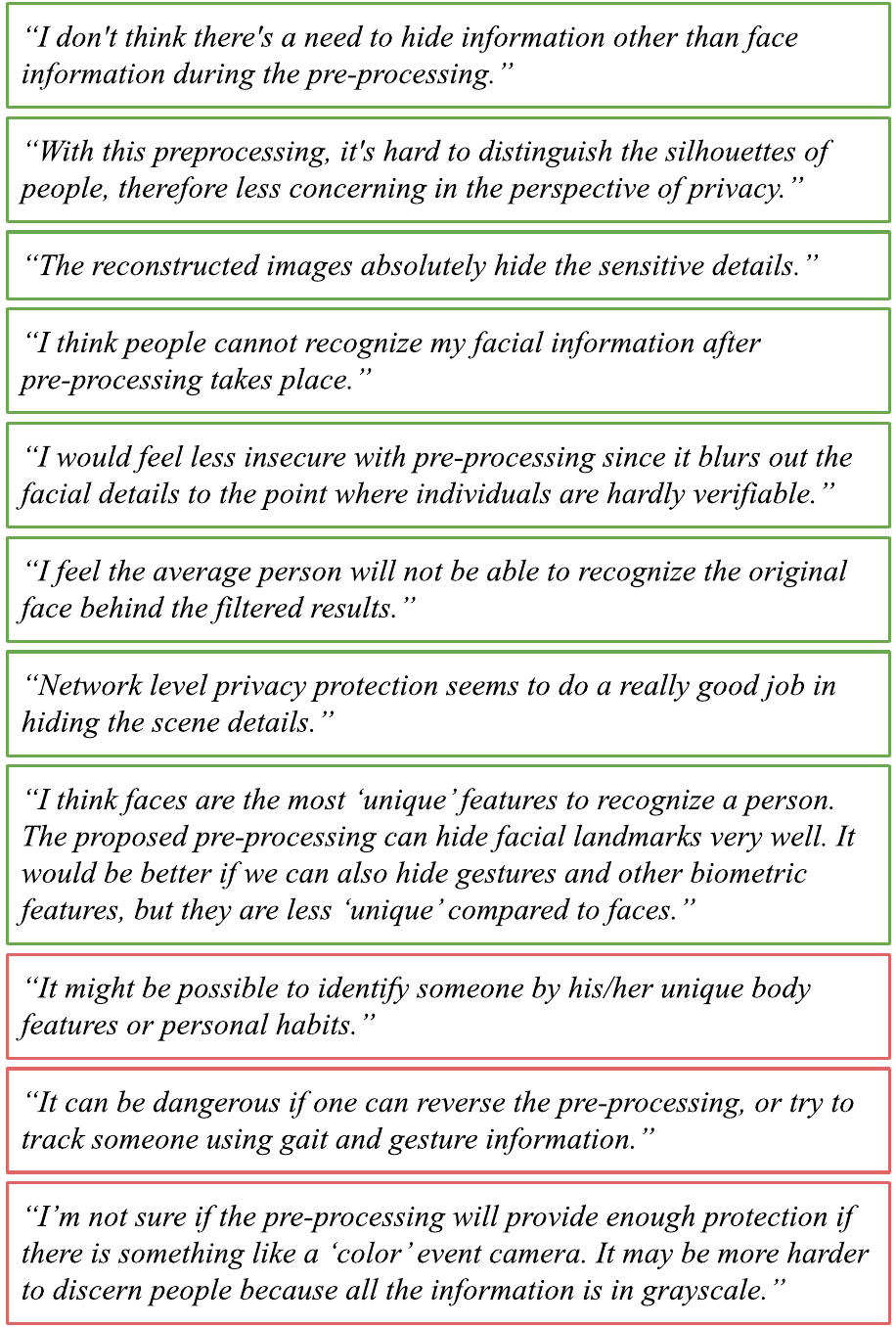}
   \caption{Positive and negative comments from the user study about our privacy protection method.}
   \label{fig:comments}
   \vspace{-1em}

\end{figure}

\subsection{User Study}
\label{sec:supp_user_study}
\subsubsection{User Comments}
Along with the insecurity scoring, we requested the users to optionally leave comments about their evaluations.
We share the user comments in Figure~\ref{fig:comments}.
Overall, the positive comments emphasize the fact that for sensor level protection the facial details are sufficiently removed and for network level protection the reconstructions successfully hide the entire user's view.
Nevertheless, there were some negative comments about the sensor level protection that other signals such as unique gestures or body features are not obscured.
Devising a privacy protection method that can cover a wider range of biometric signals is left as future work.

\subsubsection{Full Survey}
We share the full survey content used for the user study as a separate file {\small \texttt{survey.pdf}}.
Note that we showed multiple videos for each scenario in the actual user study, and for sensor level evaluation we additionally showed samples of cropped faces reconstructed after processing.

\bibliographystyle{IEEEtran}
\bibliography{main}

\begin{IEEEbiography}[{\includegraphics[width=1in,height=1.25in,clip,keepaspectratio]{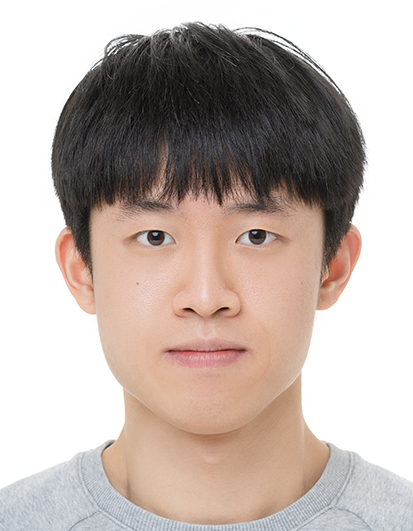}}]{Junho Kim} received the bachelor's degree from Seoul National University and the Ph.D. degree from the same university in 2025.
He is currently a postdoctoral scholar at Seoul National University.
He has published papers at prestigious venues including CVPR, ICCV, and ECCV.
His research interests include 3D reconstruction, scene understanding, and event-based vision.
\end{IEEEbiography}

\begin{IEEEbiography}[{\includegraphics[width=1in,height=1.25in,clip,keepaspectratio]{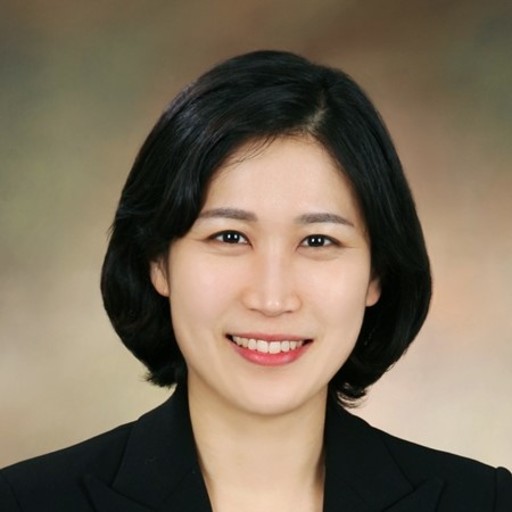}}]{Young Min Kim} is an Associate Professor in the Department of Electrical and Computer Engineering at Seoul National University, Seoul, Korea, where she is leading a 3D vision lab. She received a B.S. from Seoul National University in 2006 and an M.S. and Ph.D. in electrical engineering from Stanford University in 2008 and 2013, respectively. Before joining SNU, she was a Senior Research Scientist at the Korea Institute of Science and Technology (KIST). Her research interest lies in 3D vision, where she combines computer vision, graphics, and robotics algorithms to solve practical problems. She serves as an area chair in CVPR, ICCV, ACCV, program committee for Pacific Graphics, AAAI, and technical papers committee for SIGGRAPH Asia. She is also a program chair for 3DV 2026.
\end{IEEEbiography}

\begin{IEEEbiography}[{\includegraphics[width=1in,height=1.25in,clip,keepaspectratio]{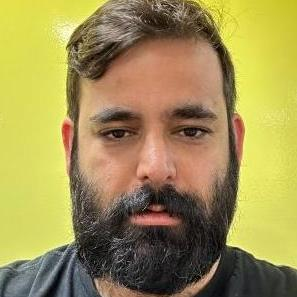}}]{Ramzi Zahreddine} received his B.S. in Optical Engineering from Rose-Hulman Institute of Technology, and his M.S. and Ph.D. in Electrical Engineering from University of Colorado Boulder. He is currently an Optical Sensing Incubation Engineer at Snap Inc. His research interests center around the intersection of computational imaging and novel optical components applied to augmented reality systems.
\end{IEEEbiography}

\begin{IEEEbiography}[{\includegraphics[width=1in,height=1.25in,clip,keepaspectratio]{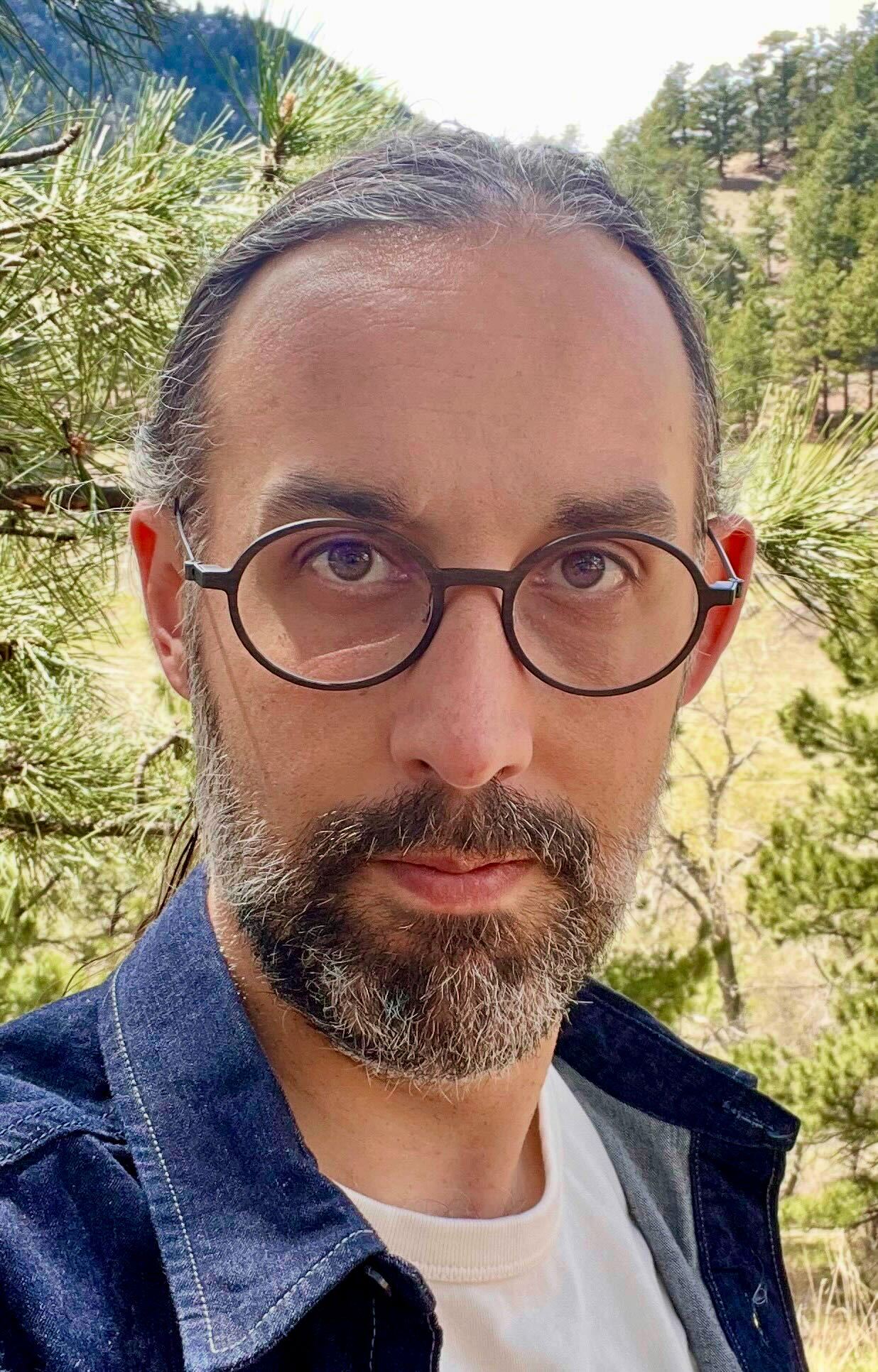}}]{Weston A. Welge} received the Ph.D. degree in optical sciences from The University of Arizona in 2016. He currently leads the camera and sensing hardware team at Snap Inc. responsible for sensing R\&D for augmented reality glasses.
\end{IEEEbiography}

\begin{IEEEbiography}[{\includegraphics[width=1in,height=1.25in,clip,keepaspectratio]{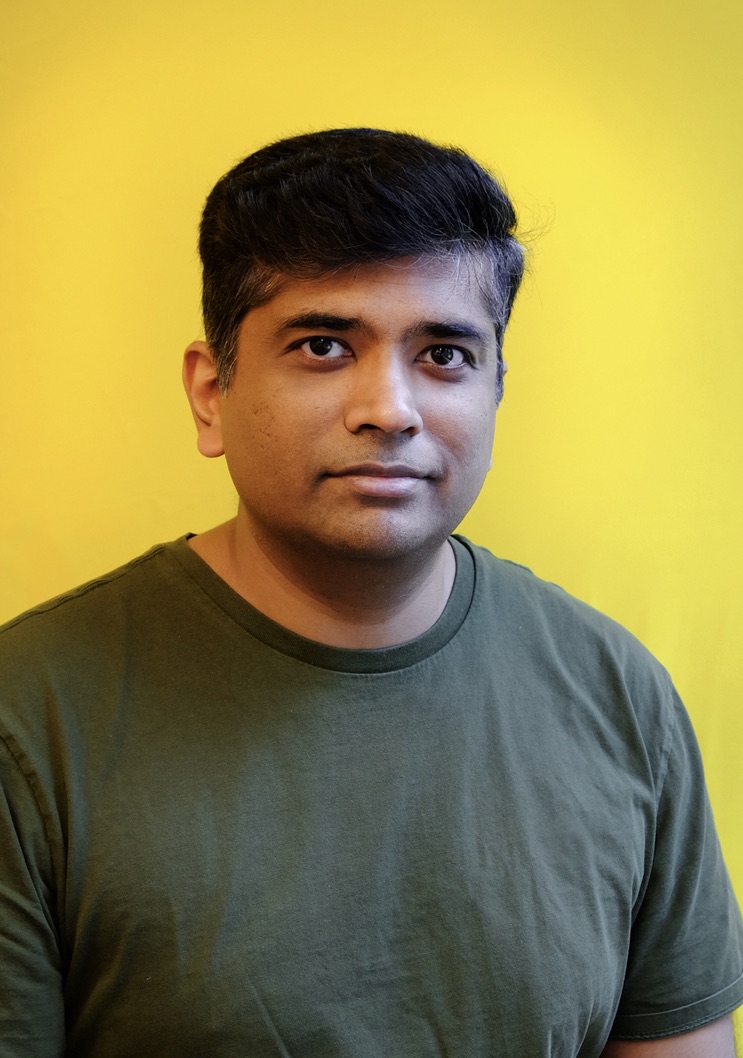}}]{Gurunandan Krishnan} is the Sr. SVP of Machine Learning at OtoNexus Medical Technologies. He holds a Masters degree in Computer Science from Columbia University. His research interests include computer vision, imaging, acoustics, augmented reality and machine learning at the edge.
\end{IEEEbiography}

\begin{IEEEbiography}[{\includegraphics[width=1in,height=1.25in,clip,keepaspectratio]{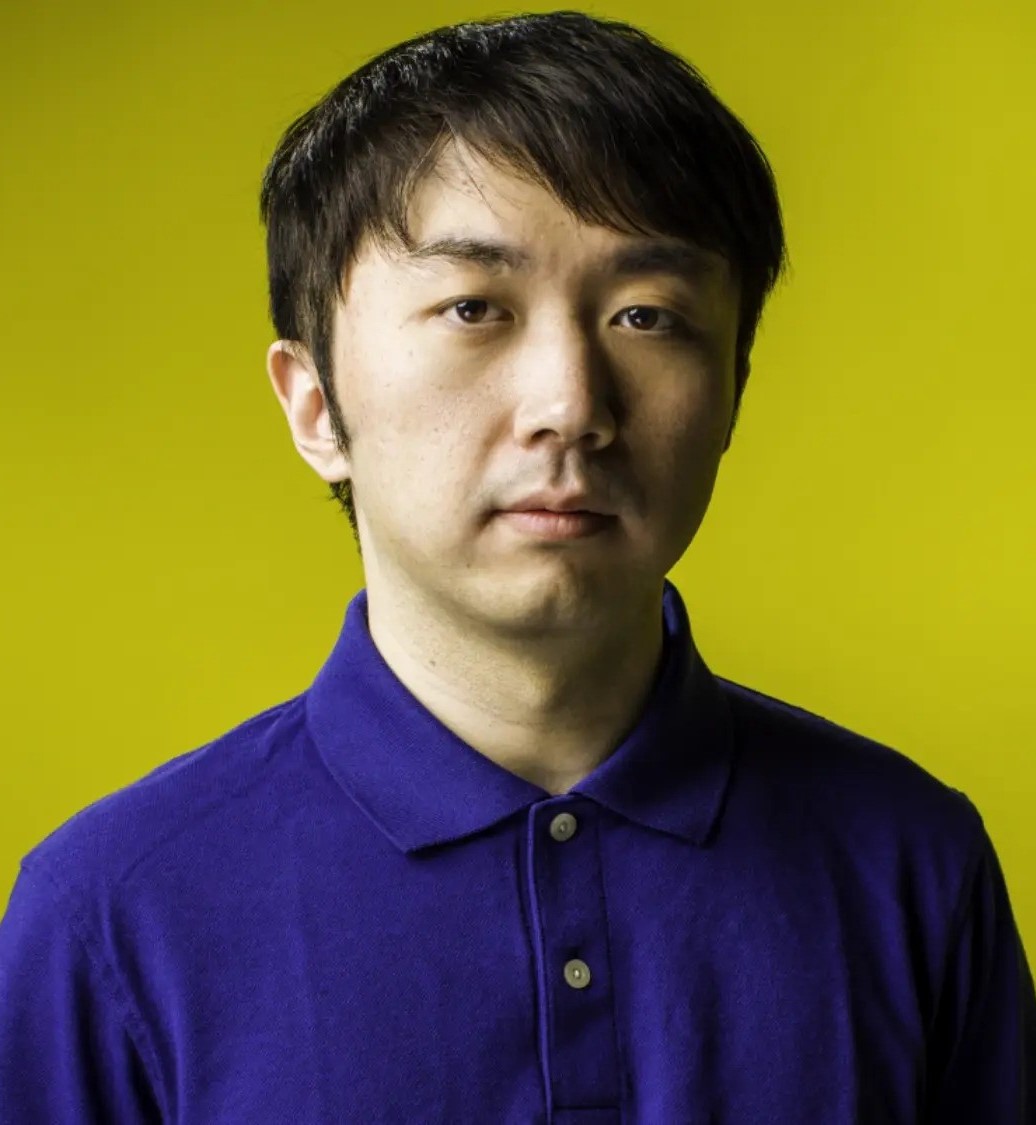}}]{Sizhuo Ma} is a Research Scientist at Snap Inc. He obtained a Ph.D. degree from University of Wisconsin-Madison in 2022. He received his bachelor's degree from Shanghai Jiao Tong University. His research interests include computational imaging, computational photography and low-level vision. He has published papers at prestigious journals and conferences including CVPR, ECCV, Siggraph, MobiCom, ISMAR, etc.
\end{IEEEbiography}

\begin{IEEEbiography}[{\includegraphics[width=1in,height=1.25in,clip,keepaspectratio]{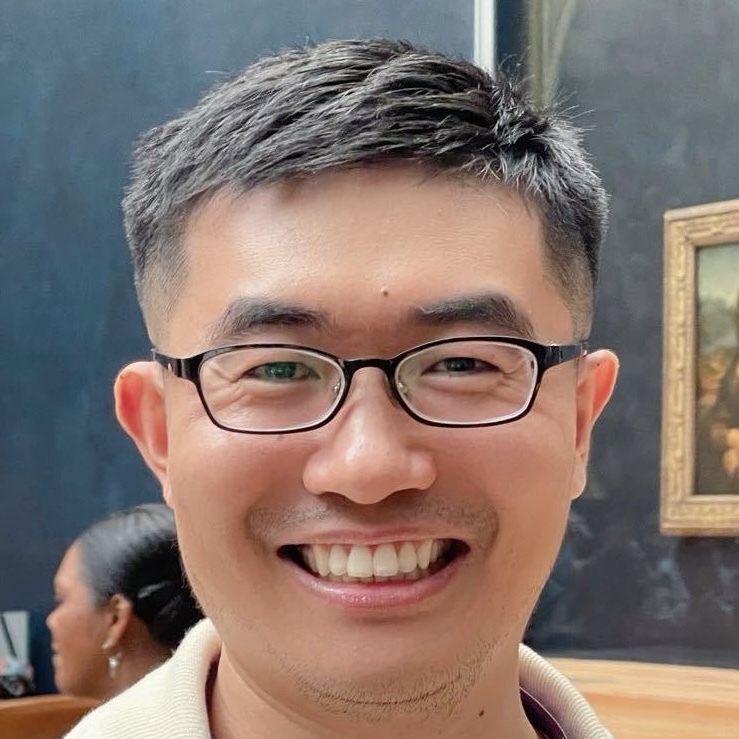}}]{Jian Wang} is a Staff Research Scientist at Snap Inc., focusing on computational photography and imaging. He has published in top-tier venues such as CVPR, MobiCom, and SIGGRAPH, and has contributed numerous features to production. He has received the best paper award from SIGGRAPH Asia, 2024, and the 4th IEEE International Workshop on Computational Cameras and Displays, and the best poster award from IEEE Conference on Computational Photography 2022. He has served as an Area Chair for CVPR, NeurIPS, ICLR, ICML, etc. Jian holds a Ph.D. from Carnegie Mellon University.
\end{IEEEbiography}

\vfill

\end{document}